\documentclass[a4paper,fleqn]{cas-dc}

\usepackage[numbers]{natbib}

\usepackage{times}
\usepackage{soul}
\usepackage{url}
\usepackage[switch]{lineno}

\hypersetup{hidelinks}
\usepackage{amsmath,amssymb,amsfonts}
\usepackage{amsthm}
\usepackage{algorithmic}
\usepackage{graphicx}
\usepackage{bm}
\usepackage{textcomp}
\usepackage{xcolor}
\usepackage{booktabs}
\usepackage{multirow}
\usepackage{enumitem}
\usepackage{hyphenat}
\usepackage{diagbox}
\usepackage{multirow}
\usepackage{tabularx}
\usepackage{makecell}
\usepackage{tcolorbox}
\usepackage{microtype}
\usepackage{xurl}
\tcbuselibrary{listings,breakable}
\tcbset{breakable, before skip=6pt, after skip=6pt}

\def\tsc#1{\csdef{#1}{\textsc{\lowercase{#1}}\xspace}}
\tsc{WGM}
\tsc{QE}
\tsc{EP}
\tsc{PMS}
\tsc{BEC}
\tsc{DE}
\ExplSyntaxOn
\cs_set:Npn \__first_footerline: { \mbox{} }
\ExplSyntaxOff

\begin{document}
\let\WriteBookmarks\relax
\raggedbottom

\shorttitle{SAGE-Loop: Reliable Closed-Loop LLM-Driven AutoML}
\shortauthors{Gu et~al.}

\title [mode = title]{SAGE-Loop: Reliable Closed-Loop LLM-Driven AutoML with Trial-and-Correction and Adaptive Ensembling}

\author{Junquan Gu}
\author{Xiaoqing Yu}
\author{Mi Wen}

% \affiliation[1]{organization={Shanghai University},
%                 addressline={99 Shangda Road},
%                 city={Shanghai},
%                 postcode={200444},
%                 country={China}}

% \cortext[cor1]{Corresponding author}

\begin{abstract}
Automated machine learning (AutoML) is reshaping data-driven science and industrial practice, and as large language models are introduced into AutoML, pipeline reliability becomes as important as automation efficiency. However, existing AutoML still struggles to realize instant feedback and adaptive optimization during execution, so once a run drifts into a suboptimal or failed state, it lacks a process-level correction mechanism. The fundamental pathology lies in its one-way pipeline: intermediate failures are typically terminated or bypassed, while fixed paradigms often strengthen model generation but leave ensemble decisions static, weakening both execution reliability and the controlled use of structural diversity. This indicates that LLM-driven AutoML needs a closed-loop ability for trial-correction-improvement together with evidence-based use of model diversity. To this end, we propose SAGE-Loop, a reliable closed-loop, self-adaptive, LLM-driven AutoML framework that performs multi-round generation and validation for trial-and-repair, and adaptively selects ensemble strategies in both supervised and unsupervised tasks, thereby unifying how to generate with how to use models. Across 20 public datasets, SAGE-Loop consistently improves performance and stability on classification, regression, and clustering tasks. Additional results show higher single-call recovery with diagnostic repair (113/240) than with generic repair (99/240).
\end{abstract}

% \includegraphics{figs/cas-grabs.pdf}

% \begin{highlights}
% \item SAGE-Loop uses prompt--execution feedback to repair model code.
% \item Validation guides adaptive selection of supervised ensemble methods.
% \item Co-association--based consensus integrates diverse clusterings.
% \end{highlights}


\maketitle
\setlength{\parfillskip}{0pt plus 0.35\linewidth}

\section{Introduction}

Automated machine learning (AutoML)~\cite{MLSYS2021_1ccc3bfa,6119056,HE2021106622,Barbudo2023-uv,JMLR:v25:22-0493,NEURIPS2020_62d75fb2,10.1007/11875581_6,pmlr-v220-roberts23a,hollmann2022tabpfn,Hollmann2025-xh} is reshaping data-driven science and industrial practice~\cite{YU2025113809,10.1145/3696410.3714567,Yu_Liu_Luo_2024,10.1145/3701716.3715587,computers10020024,9002839}, lowering the barrier to entry, shortening iteration cycles, and enhancing cross-task transfer. By automating feature handling, model generation, and configuration selection, it helps non-experts obtain reusable, portable, and interpretable pipelines under limited compute budgets~\cite{9034023,8897716,9194301,WEN2024111636}. However, as large language models are increasingly introduced into AutoML pipelines, a long-overlooked obstacle becomes more consequential: current AutoML still struggles to realize \emph{instant feedback with process-level correction} during execution, so once a run drifts into a suboptimal or failed state, recovery depends on how the system uses execution feedback.

Mainstream AutoML systems (e.g., H2O~\cite{ledell2020h2o}, Auto-sklearn~\cite{10.5555/3586589.3586850}, TPOT~\cite{olson2016tpot}) predominantly adopt a ``preset search space $\rightarrow$ one-shot solve'' paradigm, and even when training diverges, features are ineffective, or implementations error out, the generation logic rarely adapts during execution~\cite{pmlr-v64-olson_tpot_2016,DBLP:journals/corr/abs-2111-02508,NIKITIN2024112363}. Meta-learning typically serves as a warm start rather than mid-course behavioral correction~\cite{DBLP:journals/corr/abs-2012-05390}. Analysis suggests the dominant mode is ``preset the space $\rightarrow$ run once $\rightarrow$ pick the best single model or apply fixed voting,'' which systematically underutilizes structural complementarity and diversity; even when stacking is used, the level-2 learner is often a fixed template that is difficult to tailor to the current base-model ensemble and task context~\cite{10.1145/3643564}.

\begin{figure*}[pos=tb,width=\textwidth]
\centering
\includegraphics[width=\linewidth]{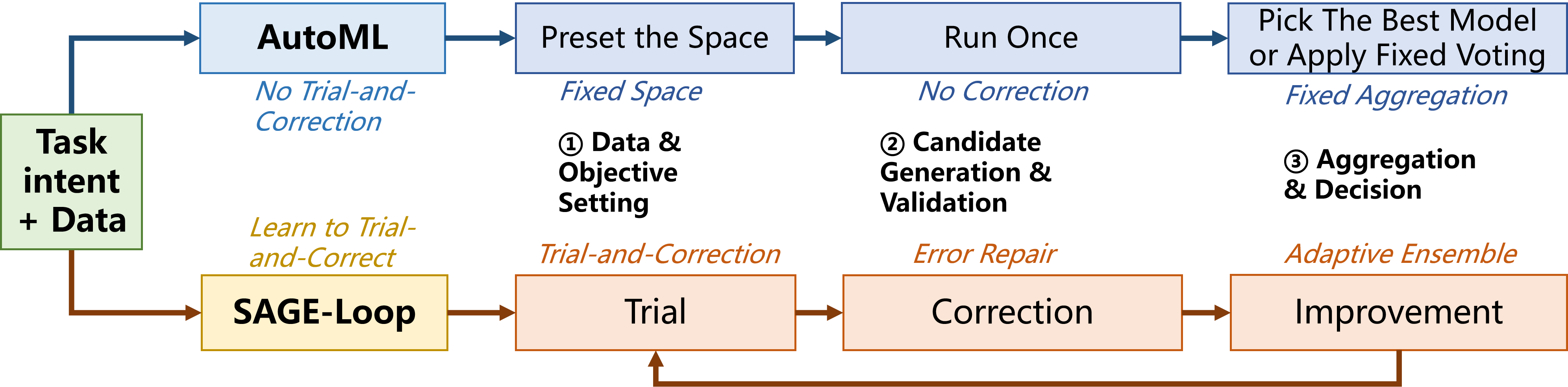}
\caption{Difference: The upper workflow uses a one-way pipeline without trial-and-correction or adaptive ensemble use, whereas SAGE-Loop introduces a closed-loop process of trial, correction, and closed-loop improvement.}
\label{fig:motivation}
\end{figure*}

Building on these observations, we identify two limitations of one-way AutoML pipelines, namely \textbf{limited use of error feedback for correction} and \textbf{limited use of evidence to combine diverse models}. These deficiencies push key stages toward unreliable or suboptimal outcomes:
1) If features are ineffective, the limitation propagates directly to downstream task performance;
2) If a model mismatches the dataset or objective, the system cannot ``improve while running'' to escape local failures;
3) If diversity of the hypothesis space is emphasized, the search cost explodes while existing models' advantages are not translated into controlled cooperative gain.
At root, current AutoML executes as a one-way pipeline; once a middle step fails, the run is typically terminated or bypassed rather than iteratively refined—there is no human-like loop of trial $\rightarrow$ correction $\rightarrow$ improvement $\rightarrow$ convergence. Moreover, fixed-form AutoML often strengthens \emph{generation} while leaving \emph{ensemble} static or absent, weakening both execution reliability and the controlled use of structural diversity, as shown in Figure~\ref{fig:motivation}.

We contend that large language models (LLMs)~\cite{rozière2024codellamaopenfoundation,touvron2023llama2openfoundation,NEURIPS2023_43e9d647,10.1145/3551349.3559548,bubeck2023sparksartificialgeneralintelligence,fang2024large} make it feasible to address reliable correction and evidence-based diversity utilization: on the one hand, \textbf{Ability to trial-and-correct}—LLMs can rewrite prompts or code based on error logs and their own structural context, forming a low-cost, training-free mechanism for lightweight trial-and-correction that immediately feeds failures and suboptimal outcomes back into generation; on the other hand, \textbf{Evidence-grounded use of diversity}—LLMs can proactively produce structurally distinct base learners and stylistically diverse clusterers, supplying the raw material for error decorrelation and structural complementarity; simultaneously, they can synthesize level-2 learners and integration logic on demand, connecting \emph{how to produce} with \emph{how to use (ensemble form)} in a task-aware manner.

Motivated by this, we propose \textbf{SAGE-Loop} (Self-Adaptive Generative Ensemble in Closed Loop), a reliable closed-loop, self-adaptive AutoML framework for tabular tasks. Its key innovation is not merely expanding the search space, \textbf{but enabling AutoML to correct itself during execution and to use model diversity in an evidence-based manner.} SAGE-Loop achieves end-to-end automation through three components:
\textbf{1) Feedbackable feature generation}: conditionally generate candidate features from task and data patterns, and form interpretable, reusable subsets via task-aware selection criteria (e.g., mutual information / F-test / variance and structural scoring);
\textbf{2) Trial-and-correction model generation}: employ multi-round prompting with execution checks, combining error-driven repair and performance revision to sustain structural diversity while correcting toward better solutions;
\textbf{3) Adaptive ensemble}: for supervised tasks, adaptively select among stacking / bagging / voting (with the stacking level-2 learner LLM-synthesized to fit the current base-model set); for unsupervised tasks, derive robust consensus over heterogeneous clusterings via co-association matrices and spectral clustering, with automatic $k$ estimation and invalid-result filtering.

We validate SAGE-Loop on 20 public datasets spanning classification, regression, and clustering, observing consistent gains in performance and stability. Ablations and visual analyses further demonstrate the robustness and interpretability benefits induced by the closed-loop trial-and-correction and principled diversity utilization. Additional tests on 125 failures show that diagnostic repair recovers 74.4\% of candidates within two attempts, compared with 64.0\% for generic repair. \noindent\textbf{Our contributions are summarized as follows:}
\begin{itemize}
    \item \textbf{Method-level contribution:} We propose a multi-round LLM-guided model synthesis method that treats the LLM as a structure search agent. This approach enables the generation of a diverse, valid, and high-quality base model pool, while supporting execution-aware repair within AutoML pipelines.
    \item \textbf{Mechanism-level contribution:} We design a unified adaptive ensemble that realizes diversity for both supervised and unsupervised tasks.
    \item \textbf{Paradigm-level contribution:} We establish a prompt--execution closed-loop workflow for reliable AutoML pipelines and assess execution recovery and held-out predictive quality after repair or regeneration.
\end{itemize}

\noindent\textbf{Code and Artifacts.}
The code, prompts, and reproduction scripts will be released upon publication.

\section{Related Works}

\subsection{LLM-Augmented AutoML Systems}

Recent advances in large language models (LLMs) have sparked growing interest in enhancing AutoML pipelines~\cite{lindauer2024positionactionhumancenteredautoml,zheng2025opencodeinterpreterintegratingcodegeneration}. Early work such as \cite{xu2024large,zhang2023automlgptautomaticmachinelearning} explores LLMs for generating ML pipeline components, followed by downstream optimization using AutoML tools. One-shot generation without feedback can leave invalid code unrepaired and make structure search fragile.

Subsequent efforts, including \cite{trirat2024automlagentmultiagentllmframework,tornede2024automl} and \cite{10.1145/3664647.3680665}, attempt to improve robustness via multi-agent collaboration or iterative prompting. Nevertheless, they are primarily confined to code-level recommendations rather than composition and repair of model structures. In the realm of feature engineering, systems like \cite{tsymbalov2024llmfeatures} demonstrate the interpretability and effectiveness of LLM-generated features, but rarely extend to full structural modeling or unify features and structures within a coherent loop. In contrast, \textbf{SAGE-Loop} embeds LLMs in a feedback loop for structure search, enabling iterative repair beyond one-shot generation.

\subsection{Structure-Aware and Adaptive Ensemble Learning}

Ensemble strategies are widely adopted in AutoML systems to boost robustness and generalization. Popular platforms such as \cite{NIPS2015_11d0e628,10.5555/3586589.3586850,SAHIN2023105732,Rimal2024-hx} typically rely on greedy ensemble selection or fixed voting schemes that can underuse structural diversity and complementarity among base models. Recent work, e.g., \cite{DBLP:journals/corr/abs-2012-05390,pmlr-v224-purucker23b}, introduces diversity metrics and synergy-aware criteria for selection, but most methods still focus on choosing or weighting given models rather than adapting the \emph{form} of the ensemble itself. \textbf{SAGE-Loop} instead adapts ensemble forms dynamically, rather than fixing selection or templates.

\subsection{Unsupervised AutoML and Clustering Fusion Paradigms}

While supervised tasks dominate AutoML research, support for unsupervised learning---especially clustering---remains limited. As noted in \cite{10.1145/3643564}, existing AutoML systems typically lack structured support for unlabeled tasks. Some initial explorations (e.g., \cite{app15126876}) run multiple clustering algorithms automatically but stop short of principled fusion.

Clustering-ensemble methods such as MCLA and CSPA~\cite{10.1162/153244303321897735} often assume homogeneous clusterers and a fixed number of clusters, which is ill-suited for LLM-generated clusterings that vary in both algorithm family and cluster count. To bridge this gap, \textbf{SAGE-Loop} consolidates diverse clusterings into a stable consensus through adaptive fusion.

Overall, while prior work has emphasized generation, \textbf{SAGE-Loop} highlights reliable closed-loop correction and adaptive use of diversity across both supervised and unsupervised AutoML under a shared feedback cycle.

\section{Preliminaries}

\subsection{Notation and Problem Setup}
We consider a tabular dataset $X\in\mathbb{R}^{n\times d}$. Supervised tasks use labels $y$; unsupervised tasks seek a clustering $c\in\{1,\ldots,k\}^n$ with $k$ clusters. We precompute train/validation/test splits and treat the test split as strictly held out: prompt updates, model selection, and ensemble decisions are based only on training/validation signals.

Let $T$ denote task metadata (e.g., task type, target, and metric), $\Sigma$ denote the input schema, and $M^{(r)}$ denote the round-$r$ memory that stores execution status, error logs, and validation outcomes. In multi-round LLM generation, the candidate set is $\mathcal{M}=\{m^{(1)},\ldots,m^{(R)}\}$ where $R$ is the number of rounds, and each model has a validation score $\mathrm{Perf}(m^{(r)})=\mathrm{score}^{(r)}$. The round-$r$ prompt and generated code are denoted by $p^{(r)}=\mathrm{Prompt}(T,\Sigma,M^{(r)})$ and $c^{(r)}=\mathrm{LLM}(p^{(r)})$. These quantities and execution logs guide prompt revision and ensemble selection.

For clustering, $B$ denotes the number of base clusterings, $c^{(b)}$ denotes the output of base clusterer $b$, and $k^{(b)}$ gives its cluster number. We use the co-association matrix $S$, where $S_{i,j}$ is the fraction of base clusterings assigning $i$ and $j$ to the same cluster; $S$ supports consensus via spectral clustering and estimating $k$ via majority vote or eigengap.

\subsection{LLM-based AutoML and Ensemble Learning}
We build a prompt from task metadata $T$, schema $\Sigma$, and round memory $M^{(r)}$, and obtain executable code from the LLM. After executing the round-$r$ code, we write feedback back into memory as
\begin{equation}
M^{(r+1)}=\mathrm{Update}\big(M^{(r)},L^{(r)}\big),
\end{equation}
where $L^{(r)}$ contains execution status, error messages, and validation scores. This update supports closed-loop trial-and-correction without reinforcement learning.

For supervised tasks, given base models $\{m^{(1)},\ldots,m^{(K)}\}$, stacking fits a level-2 learner $f_2$:
\begin{equation}
\hat{y}=f_2\big(m^{(1)}(X),\ldots,m^{(K)}(X)\big).
\end{equation}

Beyond stacking, validation scores guide selection among stacking, bagging, and voting. For stacking, the LLM synthesizes a level-2 learner matched to the task and current base models.

\subsection{Clustering Ensemble Background}
For $B$ clusterings, $c^{(b)}\in\{1,\ldots,k^{(b)}\}^n$; the co-association matrix
\begin{equation}
S_{i,j}=\frac{1}{B}\sum_{b=1}^{B}\mathbb{I}\big[c^{(b)}(i)=c^{(b)}(j)\big]
\end{equation}
captures cross-clustering similarity. We then apply spectral clustering on $S$ to obtain the consensus, and estimate $k$ either by majority vote over $\{k^{(b)}\}$ or by the Laplacian eigengap,
\begin{equation}
k=\arg\max_i (\lambda_{i+1}-\lambda_i).
\end{equation}

Classical meta-clustering (e.g., MCLA) fuses diverse partitions, while LLM-generated outputs additionally require execution and validity checks. To improve stability, we filter degenerate inputs (e.g., single-cluster outputs, empty clusters, or label-length mismatches) before consensus, so that the downstream consensus step operates on valid and comparable clustering outputs for the same set of samples.

\section{Methods}

\subsection{Overview of SAGE-Loop Framework}

This section presents the overall design of \emph{SAGE-Loop}. It unifies feature engineering, iterative model synthesis, and adaptive ensemble learning into an end-to-end pipeline with reliable closed-loop correction. The system ingests task metadata, data schema information, and a subset of raw samples. The workflow comprises three main stages: LLM-driven feature engineering, multi-round model synthesis, and ensemble strategy selection across task types. For supervised tasks, \emph{SAGE-Loop} selects the ensemble strategy by analyzing validation records of candidate models; for unsupervised tasks, it introduces a consensus-based meta-clustering mechanism that fuses heterogeneous clusterings generated by LLM-based base clusterers, thereby improving the stability and consistency of clustering results. Only candidates that pass executability checks and validation are propagated to downstream aggregation.

At the same time, \emph{SAGE-Loop} forms a lightweight \emph{trial-and-correction} loop across the above stages by using executability checks, error logs, validation-driven feedback, and prompt rewriting, enabling ``improve-while-running'' adaptivity and more reliable pipeline behavior.

\begin{figure*}[pos=tb,width=\textwidth]
\centering
\includegraphics[width=\linewidth]{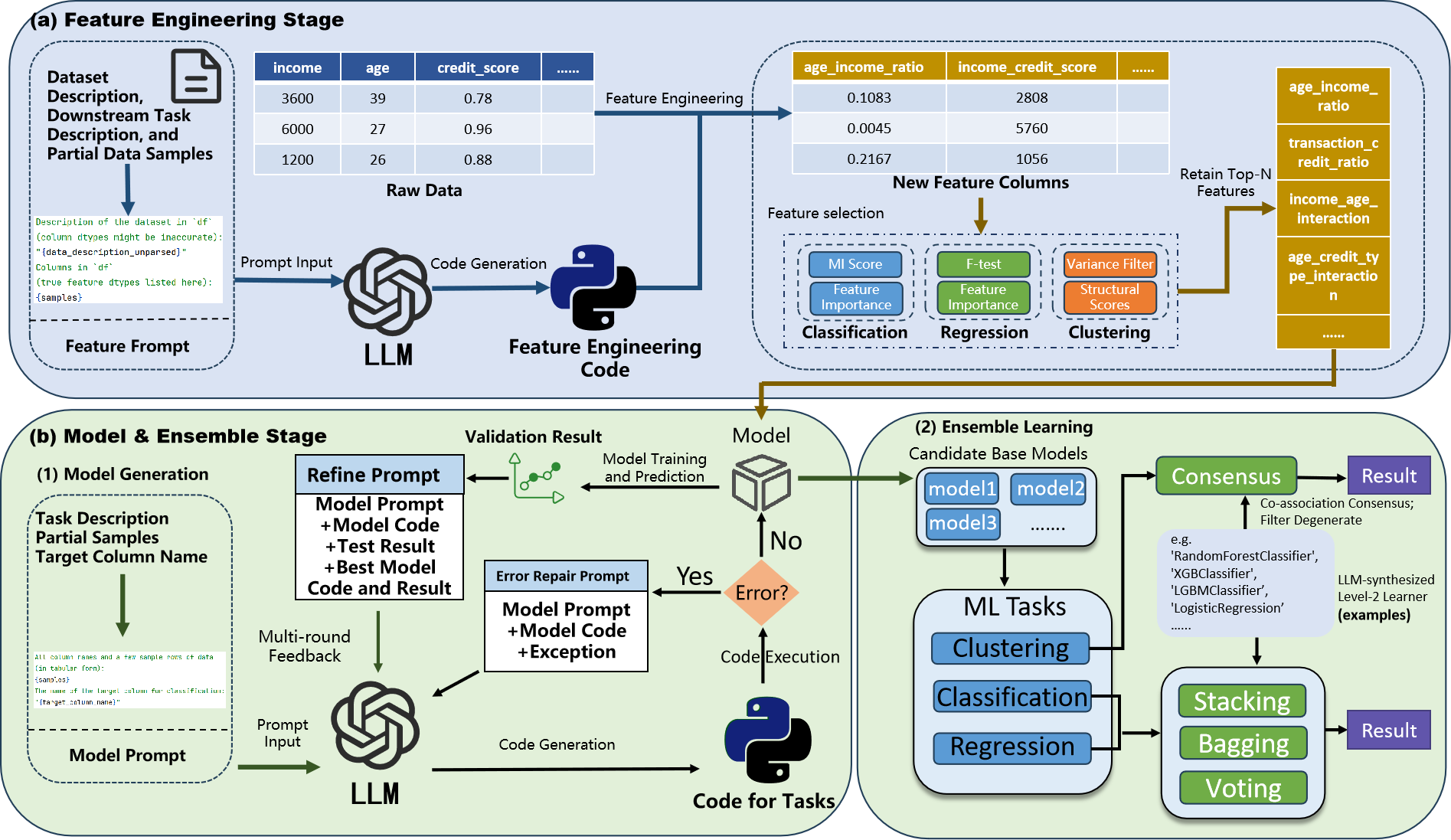}
\caption{Overview of \textbf{SAGE-Loop}. (a) The LLM synthesizes feature code from task metadata and samples, and task-aware criteria retain the top-$N$ features. (b) Multi-round model synthesis uses execution checks, error logs, validation scores, and repair prompts. Supervised tasks select stacking, bagging, or voting, with an LLM-synthesized stacking learner. Unsupervised tasks fuse valid heterogeneous base clusterings through co-association consensus and filter invalid or degenerate outputs.}
\label{fig:framework}
\end{figure*}

Figure~\ref{fig:framework} describes the pipeline, which has two stages: (a) feature engineering via prompt-guided code synthesis, and (b) iterative model generation and adaptive ensemble for downstream tasks. Each component supports automation, diversity, and robustness, and applies to both supervised and unsupervised settings. Concretely: (a) the system synthesizes feature-engineering code from prompts; (b) it performs iterative model synthesis and adaptive ensemble. While maintaining structural diversity, the components are corrected via closed-loop feedback to promote automation, robustness, execution reliability, and cross-task generality.

\subsection{LLM-Driven Feature Engineering and Multi-round Model Synthesis}

In the feature engineering stage, the system constructs a prompt from task type, data schema, and historical context:
\begin{equation}
p = \mathrm{Prompt}(T,\Sigma,M^{(r)}),
\end{equation}
where $T$ denotes task metadata, $\Sigma$ the input data schema, and $M^{(r)}$ the historical context from previous rounds. The LLM uses $p$ to generate Python code for feature engineering, producing derived features that enrich the representation. Let $F^{(r-1)}$ denote the retained feature set from earlier rounds and $\Delta F^{(r)}$ denote the newly proposed candidates in round $r$. According to the task, SAGE-Loop applies task-aware screening to retain a compact top-$N$ subset:
\begin{equation}
F^{(r)}=\mathrm{TopN}\big(F^{(r-1)}\cup \Delta F^{(r)};\, s_T(\cdot)\big),
\end{equation}
where $s_T(\cdot)$ is a task-dependent signal. For classification, mutual information and feature-importance measures are used; for regression, F-tests and feature importance are adopted; for clustering, variance-based filtering and structural scoring are employed. Later rounds retain validated features and replace ineffective or erroneous transformations.

After each round, the system records execution status, error messages, and validation scores. Error traces and context enter the next-round memory and prompt to support trial-and-correction without reinforcement-learning training.

During multi-round model synthesis, to overcome the fragility of single-shot LLM code (e.g., syntax issues, redundancy, insufficient architectural diversity), the system performs multiple rounds instead of a single architecture. In round $r$, a differentiated prompt $p^{(r)}$ is crafted to instruct the LLM to produce a \emph{different} architecture, yielding code $c^{(r)}$. The set of candidates across $R$ rounds is
\begin{equation}
\mathcal{M} = \{m^{(1)},\ldots,m^{(R)}\}.
\end{equation}
Each model must pass executability checks and is evaluated on a validation set, with performance recorded as
\begin{equation}
\mathrm{Perf}\big(m^{(r)}\big) = \mathrm{score}^{(r)}.
\end{equation}

If the generated code fails due to syntax, semantic, or runtime errors, the system constructs an \emph{error-driven repair} prompt with the error logs and feeds it back to the LLM to obtain a revised snippet; if execution succeeds but performance is unsatisfactory, a \emph{performance-driven revision} prompt is issued to guide structural updates. Together, these feedback updates preserve structural diversity while improving execution reliability and convergence.

\subsection{AutoML-Oriented Adaptive Ensemble for Supervised Tasks}

To leverage the diversity of models from multi-round generation, SAGE-Loop uses an evidence-based adaptive ensemble rather than selecting a single best model for supervised tasks. Let the selected base models be $\{m^{(1)},\ldots,m^{(K)}\}$, and let $\mathcal{A}=\{\textsc{Stacking},\textsc{Bagging},\textsc{Voting}\}$ denote the candidate strategy set. Stacking plays a central role: a level-2 learner $f_2$ is trained to integrate base predictions,
\begin{equation}
\hat{y}=f_2\big(m^{(1)}(X),\ldots,m^{(K)}(X)\big).
\end{equation}
In our framework, the level-2 learner can be generated by the LLM with additional prompting, so that its architecture is tailored to the current base-model set and task characteristics. Beyond stacking, the system also considers bagging and (weighted) voting. Using the validation records $\mathrm{Perf}(m^{(r)})$ collected in the multi-round stage, SAGE-Loop compares strategies in $\mathcal{A}$ and selects the configuration that generalizes best on the current task. This stage therefore uses feedback evidence from ``generate--check--record'' to make strategy decisions, rather than relying on a fixed ensemble template.

\subsection{AutoML-Oriented Consensus Ensemble for Unsupervised Tasks}

In unsupervised settings, LLM-generated base clusterers often differ in algorithmic family and cluster number $k$, making a single selection unstable and hard to reproduce. To convert diverse outputs into a consistent partition, the system first builds the co-association matrix $S\in\mathbb{R}^{n\times n}$:
\begin{equation}
S_{i,j}=\frac{1}{B}\sum_{b=1}^{B}\mathbb{I}\big[c^{(b)}(i)=c^{(b)}(j)\big],
\end{equation}
which gives, for samples $i$ and $j$, the fraction of base clusterings where they fall in the same cluster. The consensus number of clusters is then estimated in two ways: by \emph{majority voting} on the distribution of $\{k^{(b)}\}$ across base clusterings, and by the \emph{eigengap} heuristic on the graph Laplacian spectrum, with eigenvalues in ascending order,
\begin{equation}
k=\arg\max_i \big(\lambda_{i+1}-\lambda_i\big).
\end{equation}

After $k$ is determined, spectral clustering is applied to $S$ (as a precomputed affinity) to obtain the consensus partition. For stability and validity, base clusterings that are clearly invalid—e.g., assigning all samples to one cluster, producing empty clusters, or yielding label vectors whose length mismatches the sample size—are excluded from the consensus computation. This ``co-association + majority/eigengap + spectral + invalid-filtering'' procedure consolidates heterogeneous outputs into a stable, consistent, and more reliable clustering result, serving as the unsupervised ensemble output within the AutoML pipeline.

\begingroup
\parfillskip=0pt
Co-association is invariant to cluster-label permutations, allowing partitions with different cluster counts to contribute to a common affinity matrix without explicit label alignment.
\par
\endgroup

\section{Experiments}

To comprehensively evaluate SAGE-Loop, we formulate six research questions (RQs) aligned with the framework’s core goals and innovations. Each RQ corresponds to a specific experimental section as follows:

\begin{itemize}
    \item \textbf{RQ1 (Effectiveness across tasks):} \textit{Can SAGE-Loop outperform existing AutoML systems and LLM-based pipelines on classification, regression, and clustering tasks across diverse datasets?}  
    Section~\ref{sec:comparative_results} compares predictive performance against the baseline methods.

    \item \textbf{RQ2 (Completion recovery and execution reliability):} \textit{Can SAGE-Loop automatically repair execution failures and improve end-to-end run completion under failure-prone LLM-driven generation?}  
    Section~\ref{sec:completion_recovery} evaluates run completion and paired recovery.

    \item \textbf{RQ3 (Contribution of core components):} \textit{How much performance gain is attributable to the feature engineering and adaptive ensemble modules in SAGE-Loop?}  
    Section~\ref{sec:ablation} evaluates feature generation and ensemble learning through component ablations.

    \item \textbf{RQ4 (Mechanism adaptability and stability):} \textit{Does SAGE-Loop support task-specific mechanism design, such as optimal configuration selection in supervised tasks and automatic cluster number adaptation in unsupervised tasks?}  
    This is analyzed in Section~\ref{sec:mechanism_analysis}.

    \item \textbf{RQ5 (Interpretability and decision insights):} \textit{Can SAGE-Loop's outputs be visualized in a way that enhances interpretability and reveals prediction behavior or residual patterns?}  
    Section~\ref{sec:visualization} examines residual patterns and projected prediction surfaces.

    \item \textbf{RQ6 (Prompt-level controllability and transparency):} \textit{How diverse and responsive are SAGE-Loop’s prompts in guiding LLM-based feature and model generation, including automatic feedback handling?}  
    This is explored in Section~\ref{sec:prompt_case}.
\end{itemize}

\subsection{Experimental Setup}

For the main predictive comparisons, we report the mean and standard deviation over five random seeds for each metric. The LLM components in these comparisons use the OpenAI API with a temperature of 0.5 and a limit of 200 output tokens per call. The paired failure-recovery experiments use separate protocols, described in Section~\ref{sec:completion_recovery}. Model generation, code execution, validation, and ensemble evaluation are fully automated within a unified environment powered by an AMD Ryzen 7 8845H CPU, 32GB DDR5 RAM, and an NVIDIA GeForce RTX 4060 GPU with 8GB VRAM. Unless otherwise stated, we run a closed-loop schedule with \(R=5\) rounds per dataset. In each round, the feature generator proposes candidates and we retain the top-\(k=10\) features by the task-aware criteria; the model generator synthesizes up to 10 candidate learners, and each candidate is tuned with up to 10 hyperparameter trials. Executability checks, error logs, and validation scores trigger on-the-fly repair or replacement. No additional preprocessing is applied beyond the raw datasets as provided.

The classification comparison uses eight datasets. Five datasets, \texttt{cc1}, \texttt{cc2}, \texttt{cd2}, \texttt{cf1}, and \texttt{ld1}, are drawn from the FinBench repository~\footnote{\url{https://huggingface.co/datasets/yuweiyin/FinBench}}. The remaining classification datasets are \texttt{credit-g}~\footnote{\url{https://www.openml.org/d/31}}, \texttt{balance-scale}, and \texttt{jungle\_chess}. The ten clustering datasets are listed in Table~\ref{tab:clustering_results}. Regression evaluation uses \texttt{boston}~\footnote{\url{https://www.kaggle.com/datasets/fedesoriano/the-boston-houseprice-data}}, \texttt{concrete}~\footnote{\url{https://www.kaggle.com/datasets/maajdl/yeh-concret-data}}, \texttt{winequality}~\footnote{\url{https://www.kaggle.com/datasets/uciml/red-wine-quality-cortez-et-al-2009}}, and \texttt{california}. The datasets and protocols for the additional recovery experiments are specified in Section~\ref{sec:completion_recovery}.

All baseline methods are evaluated under their default or recommended hyperparameter settings, and every model is trained and evaluated using consistent data splits to ensure fair comparison. Guided by SAGE-Loop’s closed-loop trial-and-correction process, the clustering ensemble and the regression level-2 learner are automatically constructed and refined using signals collected in earlier rounds (executability checks, error logs, and validation scores) until validation performance converges.

\begin{table*}[pos=tb,width=\textwidth]
\centering
\scriptsize
\caption{AUC and ACC scores (\%) with standard deviations over 5 runs on 8 classification datasets. The best result in each column is shown in bold, and the second-best is underlined.}
\label{tab:classification_auc_acc_results}
\resizebox{\textwidth}{!}{
\begin{tabular}{l|cccccccc}
\toprule
\textbf{Method (AUC)} & cc1 & ld1 & credit-g & cc2 & cd2 & cf1 & balance-scale & jungle\_chess \\
\midrule
RandomForest & 65.83±0.27 & 91.51±2.54 & \underline{79.54±0.40} & 86.08±0.12 & 77.42±0.08 & 93.04±0.72 & 93.68±0.57 & 93.71±0.20 \\
XGBoost & 60.63±1.18 & 91.85±0.46 & 78.63±0.33 & 86.24±0.18 & 77.18±0.43 & 91.39±0.53 & 98.80±0.19 & 97.55±0.14 \\
LightGBM & 66.04±0.17 & 89.89±0.33 & 78.53±0.25 & \underline{86.28±0.01} & 77.70±0.08 & 91.08±0.63 & 97.66±0.29 & 96.80±0.12 \\
TPOT & 59.46±7.21 & 83.05±2.14 & 76.26±3.41 & 86.04±0.14 & 77.07±0.28 & 94.53±0.66 & 98.76±1.56 & 97.82±0.59 \\
H2O & 66.39±0.11 & 95.75±0.83 & 78.36±1.86 & 84.95±0.18 & 76.01±0.03 & 92.90±1.36 & 98.19±0.93 & 97.31±0.12 \\
AutoGluon & 65.28±0.64 & \underline{97.05±0.23} & 78.27±1.13 & 85.76±0.06 & \underline{77.72±0.04} & 92.69±0.32 & \underline{99.78±0.09} & \underline{99.65±0.04} \\
DS-Agent & 62.92±1.30 & 93.39±1.72 & 78.96±0.53 & 86.08±0.28 & 77.47±0.24 & \underline{95.15±0.60} & 97.51±0.18 & 95.94±1.27 \\
CAAFE & \underline{66.52±0.18} & 92.40±0.60 & 78.32±0.03 & 85.61±0.24 & OOM & 94.63±0.06 & 88.20±0.26 & 94.53±0.01 \\
\textbf{SAGE-Loop} & \textbf{66.98±0.09} & \textbf{97.56±0.43} & \textbf{81.08±0.24} & \textbf{87.03±0.54} & \textbf{77.91±0.13} & \textbf{97.01±1.12} & \textbf{100.00±0.00} & \textbf{99.80±0.30} \\
\bottomrule

\toprule
\textbf{Method (ACC)} & cc1 & ld1 & credit-g & cc2 & cd2 & cf1 & balance-scale & jungle\_chess \\
\midrule
RandomForest & 58.82±0.98 & 93.64±1.78 & 72.32±1.11 & 83.86±0.59 & 77.31±0.12 & 99.38±0.03 & 82.17±4.58 & 80.07±0.54 \\
XGBoost & 73.74±0.65 & 95.29±0.08 & \underline{76.40±0.72} & 87.08±0.05 & \underline{81.86±0.22} & 99.34±0.03 & 90.32±1.10 & 86.75±0.35 \\
LightGBM & 58.44±0.58 & 94.26±0.22 & 70.16±0.65 & 80.39±0.51 & 75.22±0.04 & 99.40±0.15 & 87.01±1.74 & 84.63±0.39 \\
TPOT & 77.41±0.15 & 95.03±1.13 & 75.92±0.93 & 86.89±0.19 & 81.80±0.25 & \textbf{99.54±0.06} & \underline{95.67±2.59} & 86.08±1.73 \\
H2O & 55.29±1.07 & \underline{95.90±0.35} & 72.48±2.39 & 84.54±0.59 & 78.40±0.73 & \underline{99.53±0.04} & 88.54±3.60 & 85.68±0.31 \\
AutoGluon & \underline{77.69±0.00} & 93.84±0.04 & \underline{76.40±1.43} & \underline{87.29±0.21} & 81.66±0.10 & 99.37±0.08 & \underline{95.67±1.73} & \textbf{96.19±0.33} \\
DS-Agent & 77.57±0.13 & 94.55±0.00 & 75.23±1.89 & 86.70±0.00 & 81.73±0.01 & 99.45±0.00 & 85.35±1.11 & 81.40±2.78 \\
CAAFE & 77.62±0.04 & 94.26±0.00 & 75.36±0.20 & 86.91±0.20 & OOM & 99.51±0.02 & \textbf{100.00±0.00} & 81.59±1.16 \\
\textbf{SAGE-Loop} & \textbf{77.89±0.18} & \textbf{96.63±0.12} & \textbf{79.94±0.85} & \textbf{87.32±0.05} & \textbf{82.02±0.10} & \underline{99.53±0.11} & \textbf{100.00±0.00} & \underline{92.39±4.33} \\
\bottomrule
\end{tabular}
}
\end{table*}

\begin{table*}[pos=tb,width=\textwidth]
\centering
\tiny
\caption{MAE, RMSE, and RMSLE (lower is better, mean ± std.) over 5 runs on 4 regression datasets. The SAGE-Loop results use GPT-3.5-turbo. The best result in each column is shown in bold, and the second-best is underlined.}
\label{tab:regression_results}
\resizebox{\textwidth}{!}{
\begin{tabular}{l|ccc|ccc|ccc|ccc}
\toprule
\multirow{2}{*}{\diagbox{\textbf{Method}}{\textbf{Dataset}}} & \multicolumn{3}{c|}{\textbf{boston}} & \multicolumn{3}{c|}{\textbf{concrete}} & \multicolumn{3}{c|}{\textbf{winequality}} & \multicolumn{3}{c}{\textbf{california}} \\
\cmidrule(lr){2-4} \cmidrule(lr){5-7} \cmidrule(lr){8-10} \cmidrule(lr){11-13}
& MAE & RMSE & RMSLE & MAE & RMSE & RMSLE & MAE & RMSE & RMSLE & MAE & RMSE & RMSLE \\

\midrule
RandomForest & 2.29±0.15 & 3.17±0.21 & 0.1601±0.0133 & 3.69±0.20 & 5.33±0.27 & 0.1647±0.0082 & \underline{0.42±0.02} & 0.59±0.02 & 0.0915±0.0036 & 0.33±0.01 & 0.50±0.01 & 0.1488±0.0028 \\
XGBoost & \underline{2.10±0.12} & \underline{2.89±0.13} & \underline{0.1490±0.0097} & 2.91±0.25 & 4.45±0.54 & 0.1339±0.0064 & \underline{0.42±0.02} & \underline{0.58±0.02} & \underline{0.0912±0.0033} & 0.29±0.00 & 0.44±0.01 & 0.1337±0.0017 \\
LightGBM & 2.24±0.09 & 3.11±0.04 & 0.1618±0.0087 & 3.23±0.24 & 4.74±0.44 & 0.1434±0.0065 & 0.47±0.02 & 0.62±0.02 & 0.0963±0.0037 & 0.31±0.00 & 0.46±0.01 & 0.1394±0.0027 \\
TPOT & 2.27±0.30 & 3.03±0.31 & 0.1587±0.0134 & 3.45±0.37 & 5.01±0.24 & 0.1543±0.0071 & 0.44±0.03 & 0.59±0.03 & 0.0921±0.0050 & 0.31±0.01 & 0.47±0.02 & 0.1421±0.0046 \\
H2O & 2.13±0.09 & 3.02±0.14 & 0.1497±0.0048 & 3.10±0.18 & 4.64±0.36 & 0.1407±0.0020 & \underline{0.42±0.02} & \underline{0.58±0.02} & 0.0914±0.0031 & 0.29±0.00 & 0.44±0.01 & 0.1336±0.0018 \\
AutoGluon & 2.13±0.13 & 2.97±0.19 & 0.1522±0.0089 & \underline{2.87±0.11} & \underline{4.32±0.23} & \underline{0.1285±0.0066} & 0.44±0.01 & 0.60±0.02 & 0.0927±0.0033 & \underline{0.28±0.01} & \underline{0.43±0.01} & \underline{0.1292±0.0022} \\
DS-Agent & 2.14±0.06 & 3.17±0.13 & 0.1529±0.0046 & 2.95±0.13 & 4.62±0.23 & 0.1363±0.0054 & \underline{0.42±0.01} & 0.59±0.00 & 0.0922±0.0004 & 0.33±0.08 & 0.49±0.10 & 0.1471±0.0306 \\
\textbf{SAGE-Loop} & \textbf{1.96±0.09} & \textbf{2.73±0.20} & \textbf{0.1419±0.0067} & \textbf{2.70±0.20} & \textbf{4.14±0.33} & \textbf{0.1281±0.0026} & \textbf{0.39±0.02} & \textbf{0.57±0.02} & \textbf{0.0885±0.0039} & \textbf{0.27±0.01} & \textbf{0.42±0.00} & \textbf{0.1271±0.0017} \\
\bottomrule
\end{tabular}
}
\end{table*}

\begin{table*}[pos=tb,width=\textwidth]
\centering
\tiny
\caption{ARI and NMI (\%, mean$\pm$std., 5 runs) on 10 clustering datasets. Best values are bold, second-best underlined.}
\label{tab:clustering_results}
\resizebox{\textwidth}{!}{
\begin{tabular}{l|cc|cc|cc|cc|cc}
\toprule
\multirow{2}{*}{\diagbox{\textbf{Method}}{\textbf{Dataset}}} & \multicolumn{2}{c|}{\textbf{breast}} & \multicolumn{2}{c|}{\textbf{glass}} & \multicolumn{2}{c|}{\textbf{iris}} & \multicolumn{2}{c|}{\textbf{students}} & \multicolumn{2}{c}{\textbf{seeds}} \\
\cmidrule(lr){2-3} \cmidrule(lr){4-5} \cmidrule(lr){6-7} \cmidrule(lr){8-9} \cmidrule(lr){10-11}
& ARI & NMI & ARI & NMI & ARI & NMI & ARI & NMI & ARI & NMI \\

\midrule
Kmeans & 66.74±1.12 & 55.02±1.45 & 18.50±3.59 & 33.58±5.13 & 58.07±7.74 & 64.18±2.85 & 3.34±1.79 & 13.37±1.73 & 77.80±0.96 & 73.28±0.86 \\
GaussianMixture & \textbf{77.15±0.30} & \underline{65.83±0.33} & \underline{20.70±3.11} & \underline{35.51±2.18} & \underline{82.64±15.50} & \underline{85.12±9.70} & 3.34±1.79 & 13.37±1.73 & 74.12±3.56 & 73.44±1.84 \\
AgglomerativeClustering & 57.50±0.00 & 45.69±0.00 & 13.51±0.00 & 29.55±0.00 & 61.53±0.00 & 67.55±0.00 & \underline{4.54±0.00} & \textbf{20.13±0.00} & \underline{79.70±0.00} & \underline{74.98±0.00} \\
DS-Agent & 62.12±6.49 & 51.89±2.70 & 17.34±5.17 & 31.25±5.56 & 62.01±0.00 & 65.95±0.00 & 2.87±0.00 & \underline{14.60±0.00} & 77.33±0.00 & 72.79±0.00 \\
\textbf{SAGE-Loop} & \underline{76.50±2.36} & \textbf{67.43±3.40} & \textbf{26.69±5.64} & \textbf{41.02±8.97} & \textbf{90.39±2.58} & \textbf{88.27±2.61} & \textbf{5.34±0.83} & 11.42±0.00 & \textbf{87.59±3.59} & \textbf{85.30±2.51} \\
\bottomrule

\toprule
\multirow{2}{*}{\diagbox{\textbf{Method}}{\textbf{Dataset}}} & \multicolumn{2}{c|}{\textbf{cd2}} & \multicolumn{2}{c|}{\textbf{ld1}} & \multicolumn{2}{c|}{\textbf{cd1}} & \multicolumn{2}{c|}{\textbf{ld2}} & \multicolumn{2}{c}{\textbf{cc3}} \\
\cmidrule(lr){2-3} \cmidrule(lr){4-5} \cmidrule(lr){6-7} \cmidrule(lr){8-9} \cmidrule(lr){10-11}
& ARI & NMI & ARI & NMI & ARI & NMI & ARI & NMI & ARI & NMI \\

\midrule
Kmeans & 3.25±9.90 & 2.13±4.05 & 4.52±9.22 & 1.61±3.21 & -2.53±0.00 & 0.87±0.00 & 12.39±8.33 & 6.87±2.69 & 3.47±2.08 & 6.70±3.34 \\
GaussianMixture & -3.08±0.00 & 1.65±0.00 & \underline{14.02±1.90} & \underline{5.00±0.43} & -2.40±0.00 & 0.79±0.00 & 8.12±6.09 & 2.56±0.92 & \underline{7.32±4.00} & \underline{11.56±5.77} \\
AgglomerativeClustering & -4.37±0.00 & 1.43±0.00 & -3.72±0.00 & 0.66±0.00 & -2.93±0.00 & 0.98±0.00 & \underline{16.06±0.00} & 6.33±0.00 & 7.29±0.00 & 10.09±0.00 \\
DS-Agent & \underline{16.49±5.43} & \textbf{7.76±1.75} & -0.78±2.46 & 0.22±0.08 & \underline{-0.98±1.59} & \underline{1.19±1.09} & 6.76±0.00 & \underline{7.88±0.00} & 2.56±0.00 & 8.96±0.00 \\
\textbf{SAGE-Loop} & \textbf{20.05±2.17} & \underline{4.34±2.07} & \textbf{14.03±1.16} & \textbf{5.02±0.41} & \textbf{9.41±6.08} & \textbf{2.58±1.27} & \textbf{19.12±0.83} & \textbf{9.34±3.03} & \textbf{17.47±8.13} & \textbf{13.22±5.68} \\
\bottomrule
\end{tabular}
}
\end{table*}

\subsection{Comparative Results and Analysis (RQ1)}
\label{sec:comparative_results}
We compare SAGE-Loop with traditional AutoML systems (AutoGluon~\cite{erickson2020autogluon}, H2O~\cite{ledell2020h2o}, TPOT~\cite{olson2016tpot}), tree-based models (RandomForest~\cite{Breiman2001-in}, XGBoost~\cite{chen2016xgboost}, LightGBM~\cite{ke2017lightgbm}), and recent LLM-powered methods (DS-Agent~\cite{guo2024ds}, CAAFE~\cite{hollmann2023large}), across 20 public datasets spanning classification, regression, and clustering tasks.

\paragraph{Classification.} 
Table~\ref{tab:classification_auc_acc_results} shows that SAGE-Loop consistently achieves either the best or second-best AUC and ACC scores across all eight datasets. SAGE-Loop achieves 100.00\% AUC and ACC on \textit{balance-scale} and 99.80\% AUC with an ACC of 92.39\% on \textit{jungle\_chess}. It further achieves the highest mean AUC and ACC scores on \textit{credit-g} and \textit{ld1}, and the highest mean AUC on \textit{cf1}. SAGE-Loop also achieves higher mean AUC than the compared LLM-enhanced baselines on every dataset with valid baseline outputs. On \texttt{cd2}, CAAFE terminates with an out-of-memory error, which is reported as OOM for both metrics.

\paragraph{Regression.} 
As reported in Table~\ref{tab:regression_results}, SAGE-Loop consistently achieves the lowest MAE, RMSE, and RMSLE scores across all four regression datasets. For example, on \textit{boston}, SAGE-Loop achieves an RMSE of 2.73 compared to 2.89 (XGBoost) and 2.97 (AutoGluon). RMSLE improvements are also consistent across datasets, confirming the utility of SAGE-Loop's LLM-guided feature engineering and ensemble regression mechanisms.

\paragraph{Clustering.} 
Table~\ref{tab:clustering_results} shows that SAGE-Loop achieves the highest mean score in 17 of the 20 dataset--metric comparisons across 10 clustering datasets. The remaining three comparisons are ARI on \textit{breast} and NMI on \textit{students} and \textit{cd2}, indicating that its advantages vary across datasets and metrics. On \textit{glass}, SAGE-Loop achieves ARI and NMI scores of 26.69 and 41.02, exceeding all compared baselines. On \textit{iris} and \textit{seeds}, it achieves ARI scores of 90.39 and 87.59, respectively. These results support the effectiveness of SAGE-Loop's consensus mechanism for aggregating heterogeneous LLM-generated clustering outputs.

\begin{figure}[pos=tb,width=\columnwidth]
\centering
\includegraphics[width=\linewidth]{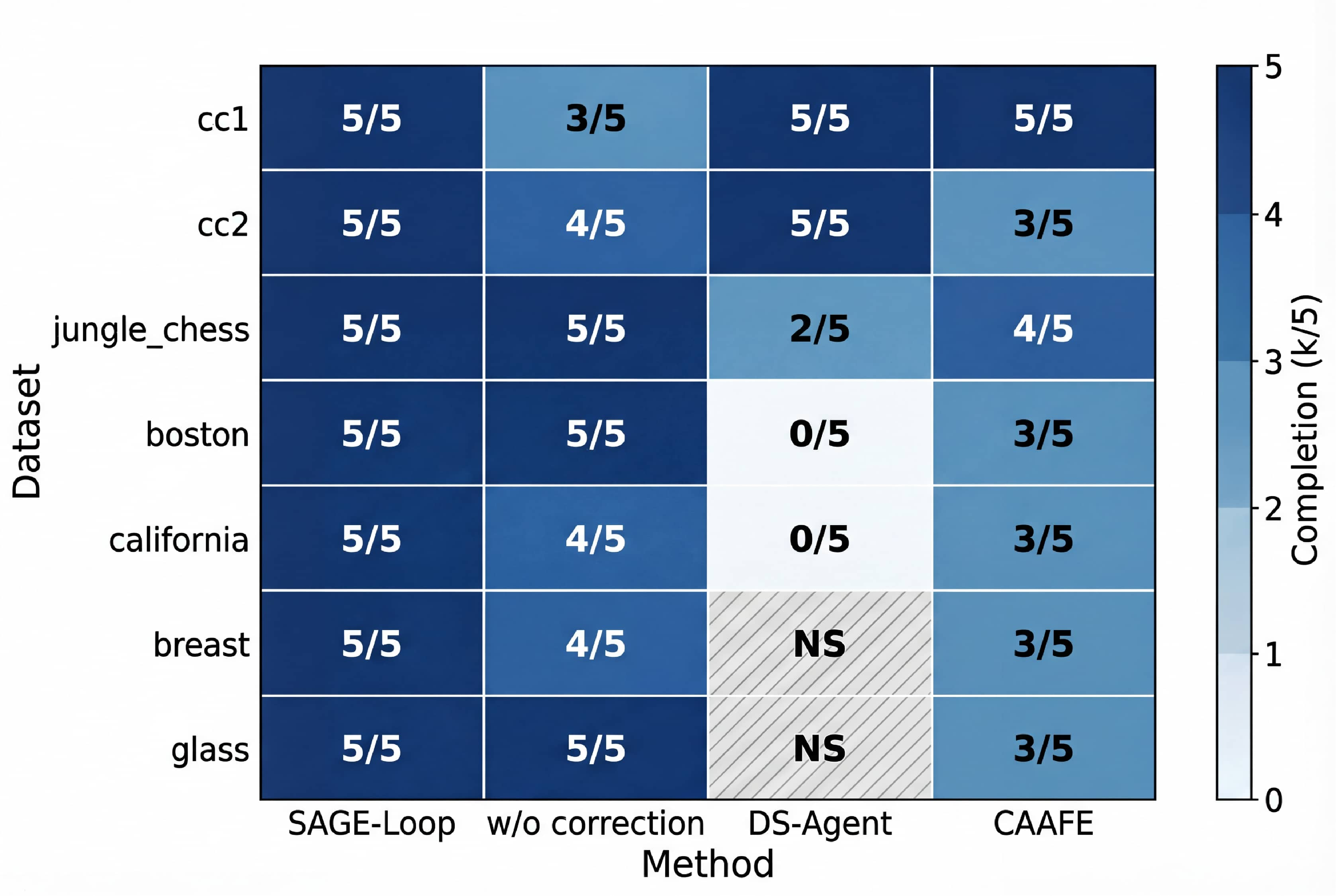}
\caption{Completion-recovery analysis on seven datasets. Cells report successful end-to-end runs out of five seeds; \textbf{NS} denotes tasks unsupported by the original configuration.}
\label{fig:completion_recovery_main}
\end{figure}

\subsection{Completion-Recovery Analysis (RQ2)}
\label{sec:completion_recovery}

To evaluate execution reliability, we measure end-to-end completion without human intervention. For the predictive performance comparisons in Tables~\ref{tab:regression_results} and~\ref{tab:clustering_results}, selected baselines required manual code repair or task-specific adaptation to produce valid outputs. SAGE-Loop required no such manual intervention. The completion analysis evaluates the baselines without these manual modifications, distinguishing their autonomous execution capability from the predictive performance obtained with manual assistance. A run is counted as successful if it completes all required stages and produces valid task outputs within the specified computational limits and stopping conditions; intermediate execution errors are permitted when handled automatically.

Figure~\ref{fig:completion_recovery_main} shows that SAGE-Loop achieves \textbf{5/5} completion on all seven datasets. Removing correction reduces completion on multiple datasets (e.g., \textit{cc1}, \textit{cc2}, \textit{california}, and \textit{breast}). Under the autonomous execution protocol, DS-Agent achieves \textbf{0/5} completion on \textit{boston} and \textit{california}, and its original configuration does not support \textit{breast} or \textit{glass}; CAAFE achieves only \textbf{3/5} completion on several datasets. SAGE-Loop supports automatic execution recovery through code repair and candidate replacement: following an execution failure, it uses error feedback to revise the failed candidate or generates a replacement candidate to continue the pipeline. Together, these recovery operations improve end-to-end completion on the evaluated tasks.

\begin{figure*}[pos=tb,width=\textwidth]
\centering
\includegraphics[width=\linewidth]{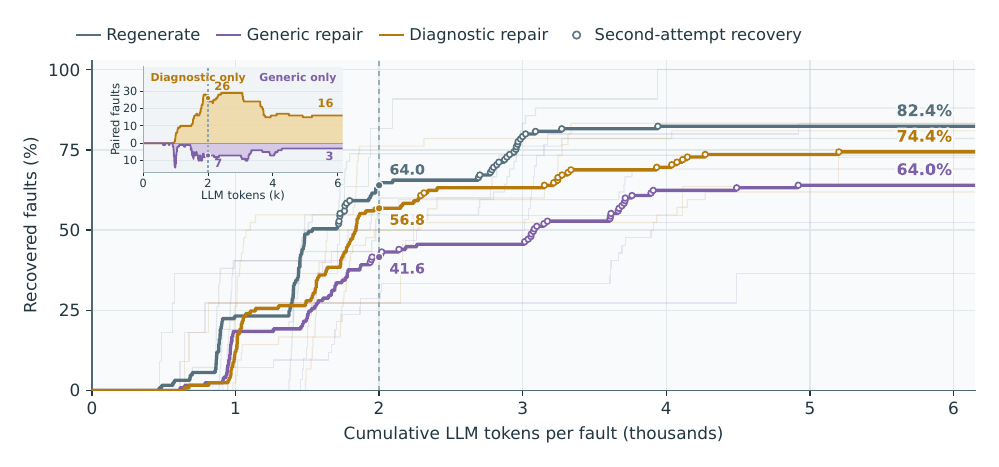}
\caption{Cumulative candidate recovery over LLM token usage on four supervised datasets. Thick steps show the pooled results over 125 failures, and thin steps show individual datasets. Open circles mark successful second attempts. The inset counts failures recovered exclusively by diagnostic repair or generic repair at each token threshold.}
\label{fig:paired_failure_recovery}
\end{figure*}

\paragraph{Paired Failure Recovery.}
To examine the effect of diagnostic feedback, we collected 125 first naturally occurring candidate failures from 192 runs on \textit{Credit Approval}, \textit{Automobile}, \textit{Seoul Bike}, and \textit{Dry Bean}. From each common failure state, we compare regeneration, repair with a generic failure prompt, and repair with a specific diagnostic prompt. Both repair strategies receive the failed code. We use GPT-4o mini (2024-07-18) with a temperature of 0.7 and a limit of 700 output tokens per call, allowing up to two attempts per strategy. A failure is counted as recovered once a candidate passes the execution and validity checks at the failed stage.

Figure~\ref{fig:paired_failure_recovery} reports recovery as a function of cumulative input and output tokens used after the initial failure. At the 2,000-token threshold, the observed recovery rates for regeneration, generic repair, and diagnostic repair are 64.0\%, 41.6\%, and 56.8\%, respectively. Diagnostic repair alone recovers 26 failures at this threshold, and generic repair alone recovers 7. Within two attempts, the three recovery rates reach 82.4\%, 64.0\%, and 74.4\%, respectively. Diagnostic repair improves recovery over generic repair by 10.4 percentage points (95\% paired bootstrap interval stratified by dataset, $[4.0,16.8]$). Relative to regeneration, its recovery rate is 8.0 percentage points lower (95\% interval, $[-16.8,0.8]$). These results quantify the contribution of specific diagnostic feedback when the repair process retains the failed code.

\paragraph{Single-Call Recovery on Held-Out Test Data.}
We evaluate 240 first natural candidate failures from \textit{Adult}, \textit{Horse Colic}, \textit{Mushroom}, \textit{Ames Housing}, \textit{Communities and Crime}, and \textit{Metro Traffic}. Each dataset contributes 40 failures in a fixed seed order. From each common failure state, we compare the four conditions in Table~\ref{tab:heldout_failure_recovery}. Each condition uses one call to GPT-4o mini (2024-07-18), with a temperature of 0.7 and a limit of 700 output tokens. The three feedback conditions share the failed code and preceding context, with exactly matched input-token counts.

\textit{Adult}, \textit{Mushroom}, and \textit{Metro Traffic} use fixed random subsets of 3,000 rows (seed 20261121), with class-stratified sampling for the two classification datasets. The other datasets retain all available rows. Approximately 20\% of rows are held out for testing. Non-temporal splits group identical predictor vectors, with linked record-ID groups additionally preserved for the \textit{Horse Colic} test split. \textit{Metro Traffic} uses the latest complete timestamps in its time-ordered subsample for testing. Test rows remain sealed until all candidate decisions are frozen. Execution recovery ($E$) requires valid execution and outputs on both development and test data. Test-qualified recovery ($Q$) additionally requires an AUC of at least $\mathrm{AUC}_{\mathrm{ref}}-0.01$ for classification or an RMSE of at most $1.02\,\mathrm{RMSE}_{\mathrm{ref}}$ for regression. Multiclass tasks use macro one-vs-rest AUC. References use logistic regression or ridge regression on the original features, with preprocessing fitted on training data. Feature repairs add the generated feature to the same reference learner. All enrolled failures remain in the denominator. The prespecified primary comparison is D versus B on execution recovery. Predictive-quality qualification is an additional endpoint. We obtain unadjusted 95\% intervals from 50,000 paired bootstrap resamples within datasets.

\begin{table*}[pos=tb,width=\textwidth]
\centering
\footnotesize
\setlength{\tabcolsep}{2.5pt}
\renewcommand{\arraystretch}{1.15}
\caption{One-call recovery of 240 paired natural failures. $E$ and $Q$ count executable and test-qualified candidates. Differences in $Q/n$ are in percentage points, with unadjusted, dataset-stratified 95\% paired bootstrap intervals for totals.}
\label{tab:heldout_failure_recovery}
\begin{tabular*}{\textwidth}{@{\extracolsep{\fill}}l*{12}{c}@{}}
\toprule
\multirow{2}{*}{Dataset} & \multirow{2}{*}{$n$}
& \multicolumn{2}{c}{Regenerate (A)}
& \multicolumn{2}{c}{Generic (B)}
& \multicolumn{2}{c}{Type only (C)}
& \multicolumn{2}{c}{Diagnostic (D)}
& \multicolumn{3}{c}{Difference in $Q/n$ (pp)} \\
\cmidrule(lr){3-4}\cmidrule(lr){5-6}\cmidrule(lr){7-8}\cmidrule(lr){9-10}\cmidrule(lr){11-13}
& & $E$ & $Q$ & $E$ & $Q$ & $E$ & $Q$ & $E$ & $Q$
& $D-B$ & $D-C$ & $D-A$ \\
\midrule
Adult                 & 40 & 18 & 8  & 13 & 9  & 17 & 8  & 21 & 9  & $0.0$   & $+2.5$  & $+2.5$  \\
Horse Colic           & 40 & 20 & 16 & 21 & 10 & 20 & 13 & 26 & 14 & $+10.0$ & $+2.5$  & $-5.0$  \\
Mushroom              & 40 & 21 & 21 & 19 & 19 & 18 & 18 & 19 & 19 & $0.0$   & $+2.5$  & $-5.0$  \\
\midrule
Ames Housing          & 40 & 14 & 13 & 15 & 13 & 14 & 10 & 13 & 11 & $-5.0$  & $+2.5$  & $-5.0$  \\
Communities and Crime & 40 & 21 & 21 & 13 & 13 & 7  & 7  & 7  & 7  & $-15.0$ & $0.0$   & $-35.0$ \\
Metro Traffic         & 40 & 30 & 29 & 18 & 18 & 18 & 18 & 27 & 27 & $+22.5$ & $+22.5$ & $-5.0$  \\
\midrule
\textbf{Total} & \textbf{240}
& 124 & 108 & 99 & 82 & 94 & 74 & 113 & 87
& \makecell[c]{$+2.1$\\$[-2.5,6.7]$}
& \makecell[c]{$+5.4$\\$[0.8,10.0]$}
& \makecell[c]{$-8.8$\\$[-15.8,-1.7]$} \\
\bottomrule
\end{tabular*}
\par\smallskip
\begin{minipage}{\textwidth}
\footnotesize
A uses the original task prompt. B, C, and D share the failed code and receive generic feedback, the exception type, or concrete diagnostics, respectively. The B/C/D prompts have equal input-token counts, and all failures remain in each arm's denominator.
\end{minipage}
\end{table*}

Table~\ref{tab:heldout_failure_recovery} reports 113 executable and 87 test-qualified candidates with concrete diagnostics, compared with 99 and 82 with generic feedback. The paired execution-recovery gain is 5.8 percentage points (95\% interval $[0.8,11.2]$). The test-qualified gain is 2.1 percentage points ($[-2.5,6.7]$).

In the secondary comparison with exception-type-only feedback, concrete diagnostics yield net increases of 19 executable and 13 test-qualified candidates. The corresponding gains are 7.9 percentage points ($[2.9,12.9]$) and 5.4 percentage points ($[0.8,10.0]$). Nine of the 13 net additional test-qualified candidates come from \textit{Metro Traffic}. Fresh regeneration yields 108 test-qualified candidates, compared with 87 for concrete diagnostics, a paired difference of $-8.8$ percentage points for D versus A ($[-15.8,-1.7]$). These results show that the observed benefit of detailed diagnostics depends on the feedback comparator and dataset. Fresh regeneration achieves the highest overall test-qualified recovery rate under this single-call protocol.

\begin{table}[pos=tb,width=\columnwidth]
\centering
\footnotesize
\setlength{\tabcolsep}{3pt}
\renewcommand{\arraystretch}{1.08}
\caption{Feature and ensemble ablations across three task types.}
\label{tab:ablation_all}

\resizebox{\linewidth}{!}{%
\begin{tabular}{lcccc}
\toprule
\multicolumn{5}{l}{\textbf{Panel A. Classification}}\\
\midrule
\multirow{2}{*}{\textbf{Setting}} & \multicolumn{2}{c}{\textbf{credit-g}} & \multicolumn{2}{c}{\textbf{cd2}} \\
\cmidrule(lr){2-3}\cmidrule(lr){4-5}
 & \textbf{ACC} & \textbf{AUC} & \textbf{ACC} & \textbf{AUC} \\
\midrule
w/o feature & 78.00$\pm$0.49 & 80.13$\pm$0.37 & 82.00$\pm$0.05 & 77.82$\pm$0.02 \\
w/o ensemble & 77.14$\pm$1.12 & 78.42$\pm$1.54 & 81.45$\pm$0.52 & 76.41$\pm$1.05 \\
\textcolor{gray}{w/o all} & \textcolor{gray}{76.11$\pm$1.33} & \textcolor{gray}{75.35$\pm$6.91} & \textcolor{gray}{81.46$\pm$0.54} & \textcolor{gray}{76.46$\pm$1.09} \\
\textbf{w all} & \textbf{79.94$\pm$0.85} & \textbf{81.08$\pm$0.24} & \textbf{82.02$\pm$0.10} & \textbf{77.91$\pm$0.13} \\
\bottomrule
\end{tabular}
}

\vspace{5pt}

\begin{tabular*}{\linewidth}{@{\extracolsep{\fill}}lccc@{}}
\toprule
\multicolumn{4}{l}{\textbf{Panel B. Regression}}\\
\midrule
\multicolumn{4}{l}{\textbf{boston}}\\
\textbf{Setting} & \textbf{MAE} & \textbf{RMSE} & \textbf{RMSLE} \\
\midrule
w/o feature & 2.00$\pm$0.07 & 2.82$\pm$0.17 & 0.1437$\pm$0.0087 \\
w/o ensemble & 2.15$\pm$0.24 & 3.17$\pm$0.36 & 0.1510$\pm$0.0145 \\
\textcolor{gray}{w/o all} & \textcolor{gray}{2.18$\pm$0.26} & \textcolor{gray}{3.36$\pm$0.32} & \textcolor{gray}{0.1577$\pm$0.0150} \\
\textbf{w all} & \textbf{1.96$\pm$0.09} & \textbf{2.73$\pm$0.20} & \textbf{0.1419$\pm$0.0067} \\
\midrule
\multicolumn{4}{l}{\textbf{concrete}}\\
\textbf{Setting} & \textbf{MAE} & \textbf{RMSE} & \textbf{RMSLE} \\
\midrule
w/o feature & 2.90$\pm$0.20 & 4.43$\pm$0.40 & 0.1372$\pm$0.0053 \\
w/o ensemble & 3.39$\pm$0.45 & 4.91$\pm$0.47 & 0.1504$\pm$0.0156 \\
\textcolor{gray}{w/o all} & \textcolor{gray}{3.81$\pm$1.12} & \textcolor{gray}{5.38$\pm$1.08} & \textcolor{gray}{0.1714$\pm$0.0516} \\
\textbf{w all} & \textbf{2.70$\pm$0.20} & \textbf{4.14$\pm$0.33} & \textbf{0.1281$\pm$0.0026} \\
\bottomrule
\end{tabular*}

\vspace{5pt}

\resizebox{\linewidth}{!}{%
\begin{tabular}{lcccc}
\toprule
\multicolumn{5}{l}{\textbf{Panel C. Clustering}}\\
\midrule
\multirow{2}{*}{\textbf{Setting}} & \multicolumn{2}{c}{\textbf{glass}} & \multicolumn{2}{c}{\textbf{seeds}} \\
\cmidrule(lr){2-3}\cmidrule(lr){4-5}
 & \textbf{ARI} & \textbf{NMI} & \textbf{ARI} & \textbf{NMI} \\
\midrule
w/o feature & 24.22$\pm$1.53 & 40.05$\pm$2.49 & 64.96$\pm$7.70 & 68.03$\pm$4.95 \\
w/o ensemble & 20.20$\pm$5.99 & 33.59$\pm$7.35 & 75.34$\pm$5.37 & 74.12$\pm$4.20 \\
\textcolor{gray}{w/o all} & \textcolor{gray}{21.65$\pm$5.83} & \textcolor{gray}{39.80$\pm$3.75} & \textcolor{gray}{64.27$\pm$15.41} & \textcolor{gray}{68.38$\pm$8.34} \\
\textbf{w all} & \textbf{26.69$\pm$5.64} & \textbf{41.02$\pm$8.97} & \textbf{87.59$\pm$3.59} & \textbf{85.30$\pm$2.51} \\
\bottomrule
\end{tabular}
}
\end{table}

\subsection{Ablation Studies (RQ3)}
\label{sec:ablation}
To assess the contribution of SAGE-Loop's core components, we perform ablation studies by selectively disabling (1) feature engineering, (2) ensemble learning, and (3) both. The results across classification, regression, and clustering tasks are summarized in Table~\ref{tab:ablation_all}. These comparisons quantify changes in task performance when each component is removed. Section~\ref{sec:completion_recovery} separately evaluates autonomous completion and paired candidate recovery.

\paragraph{Classification.} 
Removing either the feature or ensemble module degrades performance. On \textit{credit-g}, AUC drops from \textbf{81.08} to \textbf{80.13} without feature generation and to \textbf{78.42} without ensemble learning; the full model also yields the best ACC. These results indicate that both modules contribute to stable classification gains, with the ensemble component playing the larger role on this task.

\paragraph{Regression.} 
On \textit{boston} and \textit{concrete}, disabling the ensemble mechanism increases RMSE from \textbf{2.73} to \textbf{3.17} and from \textbf{4.14} to \textbf{4.91}, respectively, while removing all modules leads to the highest errors across all metrics. These results support the joint contribution of LLM-generated features and adaptive ensemble regression on the two evaluated datasets.

\paragraph{Clustering.}
On \textit{glass} and \textit{seeds}, removing either the feature or ensemble module causes clear ARI/NMI degradation. For example, on \textit{seeds}, ARI drops from \textbf{87.59} to \textbf{64.96} without feature generation and to \textbf{75.34} without ensemble learning. This confirms that both modules are crucial for stabilizing heterogeneous clustering outputs and converting diversity into reliable consensus quality.

\begin{figure}[pos=tb,width=\columnwidth]
\centering
\includegraphics[width=\linewidth]{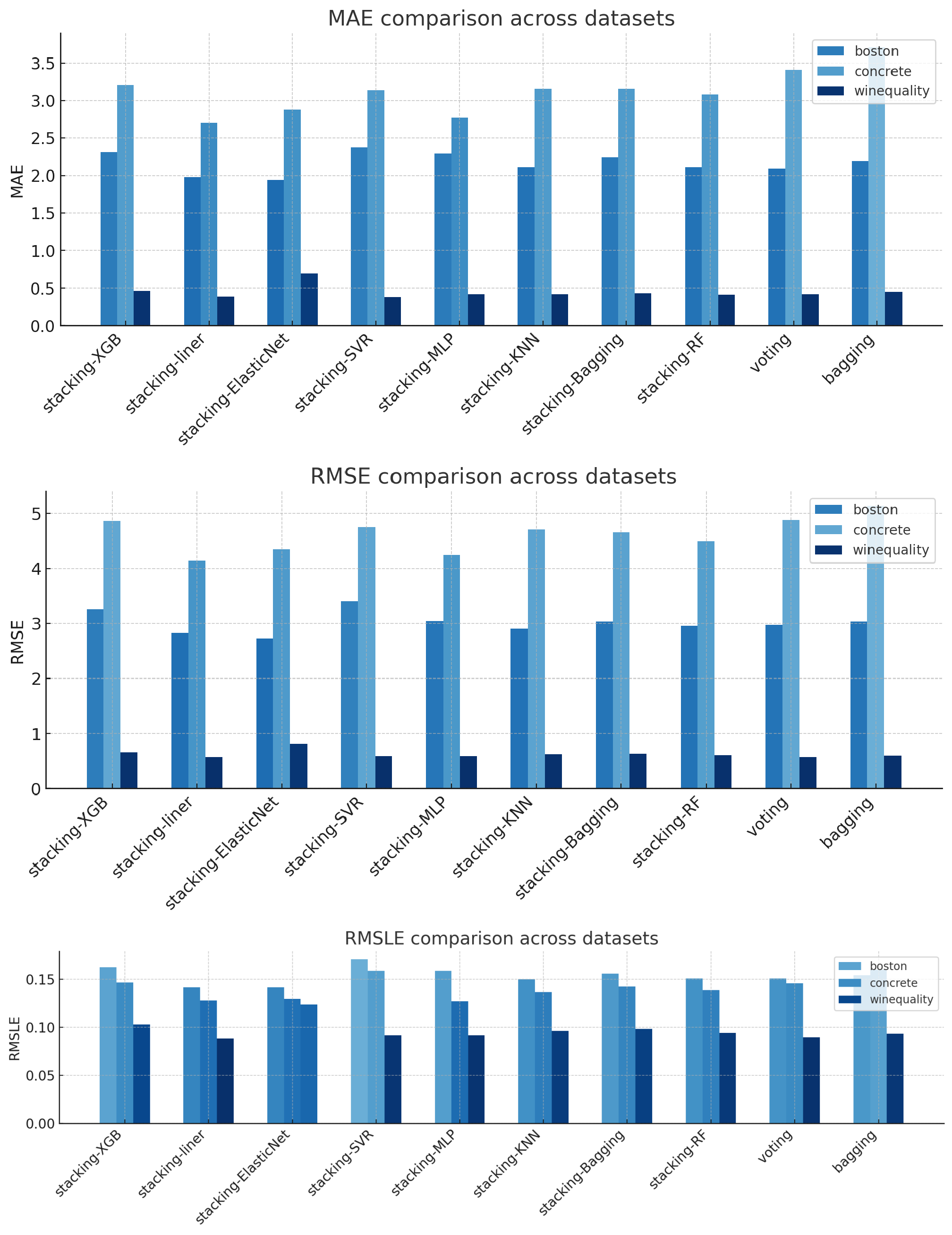}
\caption{Supervised regression: MAE, RMSE, and RMSLE across stacking variants and ensemble baselines on \textit{boston}, \textit{concrete}, and \textit{winequality} (lower is better).}
\label{fig:regression_mechanism_main}
\end{figure}
\par\smallskip

\begin{figure*}[pos=tb,width=\textwidth]
\centering
\includegraphics[width=\linewidth]{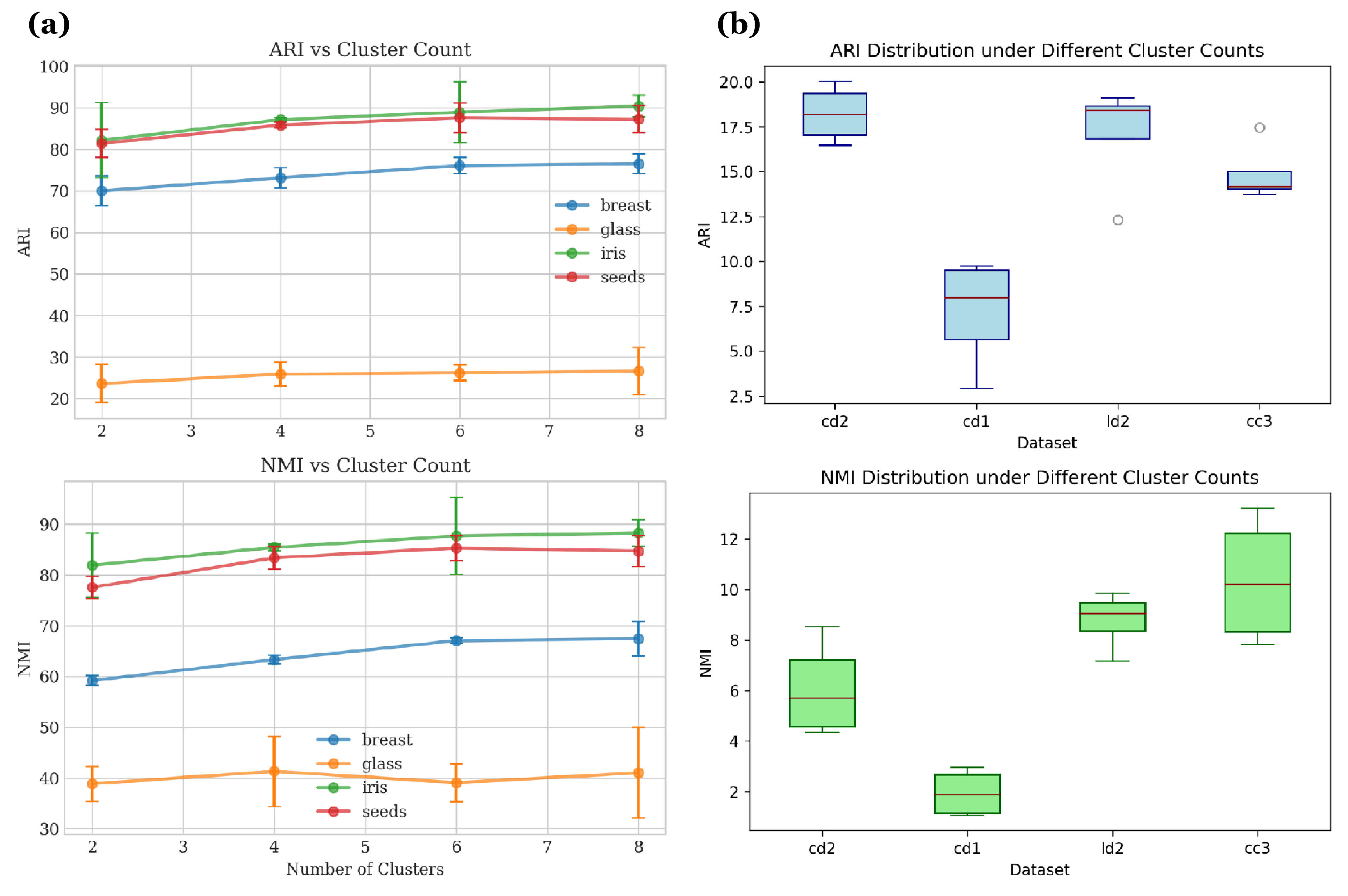}
\caption{Clustering mechanism analysis. (a) ARI and NMI under varying cluster counts on \textit{breast}, \textit{glass}, \textit{iris}, and \textit{seeds}. Error bars show mean$\pm$std over 5 runs. (b) ARI and NMI distributions under varying cluster counts on \textit{cd2}, \textit{cd1}, \textit{ld2}, and \textit{cc3}.}
\label{fig:cluster_ensemble_main}
\end{figure*}
\par\smallskip

\subsection{Mechanism Analysis (RQ4)}
\label{sec:mechanism_analysis}

We analyze how the proposed mechanisms operate in practice for both supervised and unsupervised settings using the integrated figures in the main text.

\paragraph{Supervised Regression Mechanism.}
Figure~\ref{fig:regression_mechanism_main} compares stacking variants and ensemble baselines on \textbf{MAE}, \textbf{RMSE}, and \textbf{RMSLE} across three regression datasets. Different level-2 learners reach distinct optima on different datasets, while the validation-driven loop consistently favors the configuration that best matches the current base-model pool. In particular, stacking variants are often preferable when the selected level-2 learner can exploit dataset-specific structure, whereas bagging and voting remain competitive but less consistently optimal. This behavior supports the design principle of \emph{adaptive ensemble grounded in closed-loop evidence}: rather than fixing a single meta-learner, SAGE-Loop uses validation signals to determine how model diversity should be aggregated.

\paragraph{Unsupervised Clustering Mechanism.}
Figure~\ref{fig:cluster_ensemble_main}(a) reports both \textbf{ARI} and \textbf{NMI} under varying cluster numbers. The results reveal two properties of the consensus layer: (i) \emph{automatic $k$ adaptation} via majority-vote/eigengap; and (ii) robustness to heterogeneous base clusterings, aided by degenerate filtering before fusion. On structured datasets such as \textit{iris} and \textit{seeds}, both metrics quickly reach stable high-quality regions as $k$ varies, while on more irregular datasets such as \textit{glass}, the curves remain smooth without collapse. These patterns indicate that the consensus mechanism converts diversity into stable clustering quality rather than brittle single-choice outputs.

\paragraph{Cluster Count Sensitivity across Datasets.}
Figure~\ref{fig:cluster_ensemble_main}(b) summarizes \textbf{ARI} and \textbf{NMI} distributions across datasets under cluster counts 2--8. Narrow interquartile ranges with high medians indicate strong robustness, whereas broader spreads suggest higher sensitivity to partition granularity. Even in relatively sensitive cases such as \textit{cd1}, the consensus output avoids extreme collapses, which is consistent with the degenerate-filtering step and the $k$-estimation safeguards.

\subsection{Visualization and Interpretability (RQ5)}
\label{sec:visualization}

We visualize feature enhancement on \textit{boston}, \textit{california}, \textit{concrete}, and \textit{insurance} using residual plots and PCA-projected prediction surfaces. The two representative datasets from the main discussion appear first, followed by the additional datasets, so that changes in error patterns and predicted responses can be inspected across feature representations.

\begin{figure*}[pos=tb,width=\textwidth]
\centering
\includegraphics[width=\linewidth]{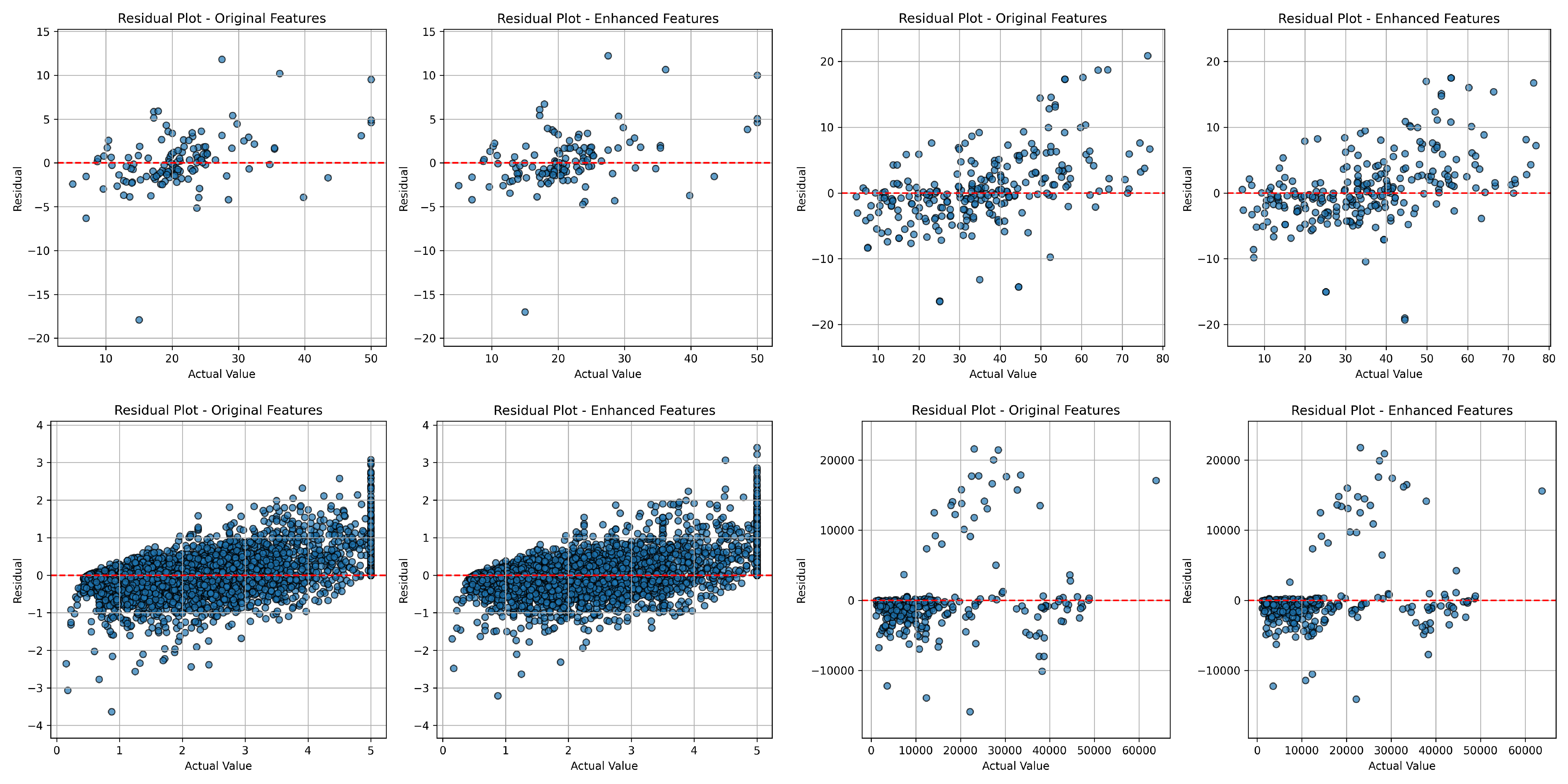}
\caption{Residual plots on four regression datasets. The left half shows the two representative datasets used in the main analysis (\textit{boston} and \textit{california}), and the right half shows two additional datasets (\textit{concrete} and \textit{insurance}). For each dataset, original features are shown on the left and enhanced features on the right.}
\label{fig:residual_main}
\end{figure*}

\paragraph{Residual Distribution.}
Figure~\ref{fig:residual_main} compares residual patterns before and after feature enhancement on four regression datasets. Residual dispersion varies with the dataset and target range. Large positive and negative residuals remain in both feature settings on \textit{concrete} and \textit{insurance}. The plots reveal dataset-dependent changes in error patterns and complement the aggregate regression metrics.

\begin{figure}[pos=!htbp,width=\columnwidth]
\centering
\includegraphics[width=\linewidth]{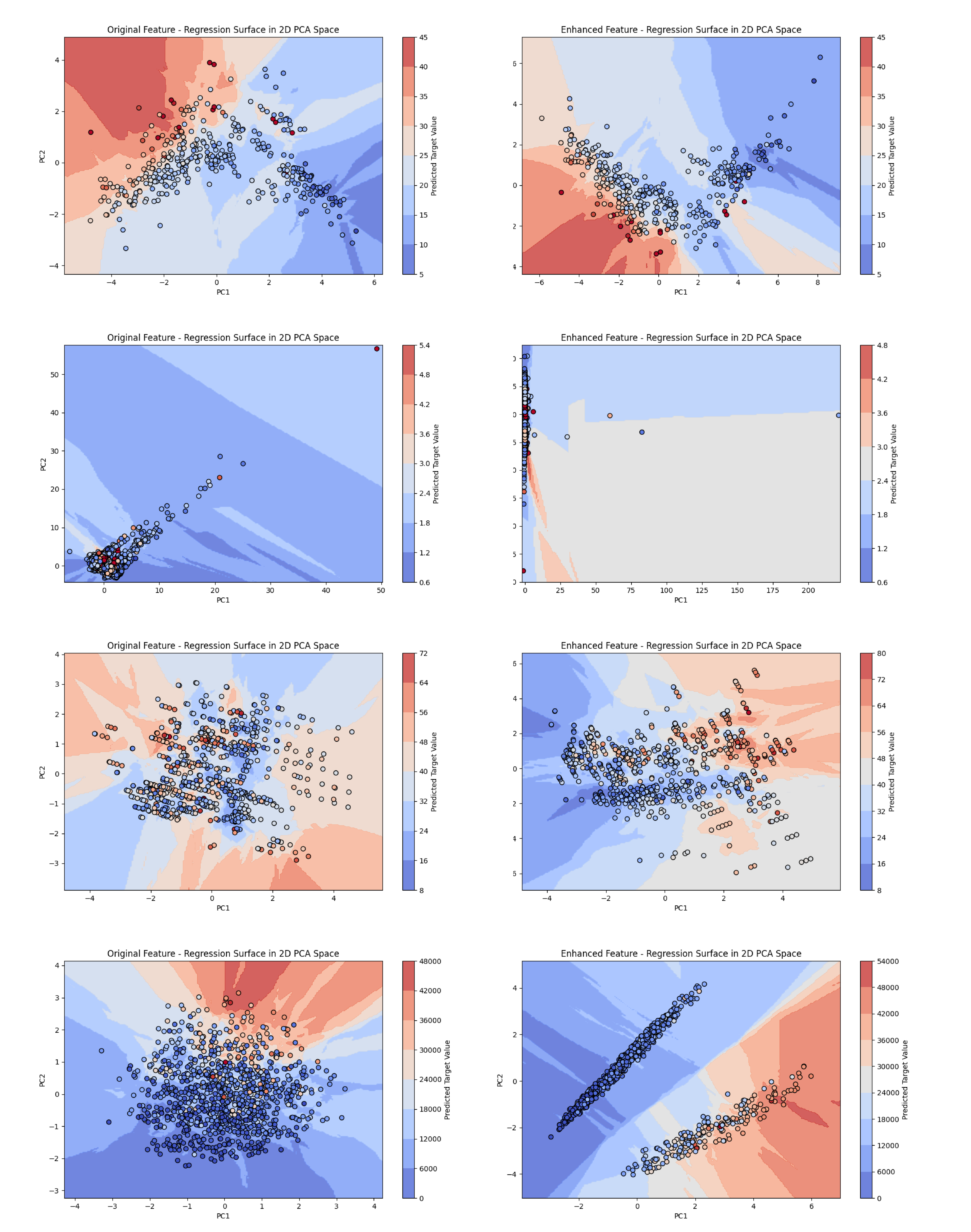}
\caption{PCA-projected regression surfaces for \textit{boston}, \textit{california}, \textit{concrete}, and \textit{insurance} (top to bottom). Original and enhanced features appear left and right.}
\label{fig:surface_main}
\end{figure}

\paragraph{Prediction Surface in PCA Space.}
Figure~\ref{fig:surface_main} visualizes predicted responses over two-dimensional PCA projections of the original and enhanced feature spaces. The panels show dataset-specific prediction patterns, with different coordinate ranges and color scales. These visualizations provide qualitative views of model behavior in each feature representation. Quantitative predictive performance is assessed separately in Section~\ref{sec:comparative_results}.

\par\smallskip

\begingroup
\setlength{\parfillskip}{0pt plus 1fil}
\subsection{Prompt Design Case Studies (RQ6)}
\label{sec:prompt_case}

To illustrate how SAGE-Loop interacts with large language models (LLMs) in a structured and iterative manner, we showcase four representative prompt templates used throughout the system pipeline: feature generation, model generation, feedback-based refinement, and error correction. Each prompt is designed to be task-specific, output-constrained, and verifiable, ensuring reliable integration between LLM reasoning and programmatic execution.

\paragraph{Feature Generation Prompt.}  
The prompt instructs the LLM to propose new interpretable numeric features based on domain-specific tabular data. It specifies strict formatting rules and demands valid Python code with embedded use-case explanation and sample data, aiming to ensure both usability and interpretability in downstream tasks.

% \clearpage
\makeatletter
\global\@colroom=\@colht
\global\vsize=\@colht
\makeatother
\begin{tcolorbox}[title={Prompt-FG (Classification Task)}, colback=gray!5, colframe=black!50, fonttitle=\bfseries]
\footnotesize
\texttt{The dataframe \texttt{cd1} is loaded in memory for a classification task.}\\
Target: ``class''.\\
Description: ``Financial client records with loan type, age, education, income, credit score, etc. Predict default status.''\\
Sampled instances: \texttt{df.head(10)}\\
Number of samples: 2738\\[4pt]

Generate one useful numeric column using transformations, combinations, or aggregations. Follow these rules:\\
- Use only existing columns\\
- Must be numeric, interpretable, and non-redundant\\
- Enhance downstream accuracy\\[4pt]

\textbf{Output format (strict):}\\
\texttt{```python}\\
\# Usefulness: (Rationale for classifying ``class'')\\
\# Input samples: 'col1': [...], 'col2': [...], ...\\
df['NEW\_COLUMN'] = transformation\\
\texttt{```end}
\end{tcolorbox}

\paragraph{Model Generation Prompt.}  
This prompt guides the LLM to generate a complete Python classifier named \texttt{myclassifier} for a binary classification task. It includes the goal (maximize AUC), allowed APIs, method constraints, and required outputs.

\begin{tcolorbox}[title={Prompt-MG (Model Generator)}, colback=gray!5, colframe=black!50, fonttitle=\bfseries]
\footnotesize
\texttt{You are a top-level machine learning expert. Your task is to propose a new classifier named \texttt{myclassifier} to maximize AUC.}\\
The model must differ from all previous ones by type or internal structure.\\

\textbf{Output must:}\\
- Be valid Python code only\\
- Include all imports\\
- Implement: \texttt{fit}, \texttt{predict}, \texttt{predict\_proba}\\
- Avoid BaggingClassifier\\
- Support AUC-oriented probabilistic output\\
- Use inputs: \texttt{train\_aug\_x}, \texttt{train\_aug\_y}, \texttt{test\_aug\_x}
\end{tcolorbox}

\paragraph{Feedback-Driven Prompt.}  
This prompt is triggered after a model is evaluated. It provides the current and best AUC so far, and asks the LLM to improve upon the last attempt by proposing a better model variant, effectively simulating an iterative optimization process.

\begin{tcolorbox}[title={Prompt-FB (Feedback Refinement)}, colback=gray!5, colframe=black!50, fonttitle=\bfseries]
\footnotesize
\texttt{Current model AUC: 0.7241}\\
\texttt{Best historical AUC so far: 0.7559}\\[4pt]
Current model code\\
\texttt{```python}\\
... (previous invalid model code) ...\\
\texttt{```}\\[4pt]
Please now propose a new classifier that is more likely to improve the AUC.\\
The model must differ from all previous ones by type or structure.\\
Only return Python code defining class \texttt{myclassifier} with the required methods.
\end{tcolorbox}

\paragraph{Error Correction Prompt.}  
If the generated code fails at runtime, this prompt automatically injects the exception message and re-asks the LLM to revise its answer, while reminding it of format restrictions, thereby reducing repetitive manual debugging efforts.

\begin{tcolorbox}[title={Prompt-ER (Error Repair)}, colback=gray!5, colframe=black!50, fonttitle=\bfseries]
\footnotesize
\texttt{The classifier code execution failed with error: \texttt{TypeError: unexpected keyword argument 'learning\_rate'}}\\
Code:\\
\texttt{```python}\\
... (previous invalid model code) ...\\
\texttt{```}\\[4pt]
Remember: your answer must only be valid Python code.\\
No explanations or extra output. Generate a corrected version now.
\end{tcolorbox}

\endgroup
\section{Conclusion}
We presented \textbf{SAGE-Loop}, a reliable closed-loop LLM-driven AutoML framework for tabular learning built on two principles: trial--and--correction during execution and adaptive ensemble based on validation evidence. Rather than merely enlarging the search space, SAGE-Loop closes the generation loop through feedbackable feature synthesis and multi-round model synthesis, and then converts structural diversity into stable gains via an adaptive ensemble layer. Experiments on 20 public datasets covering classification, regression, and clustering show consistent improvements in both performance and stability, while the completion-recovery analysis further demonstrates its ability to automatically repair execution failures and achieve reliable end-to-end run completion. Looking forward, we plan to extend SAGE-Loop to multimodal settings, study theoretical properties of closed-loop modeling such as convergence and consistency, and further optimize efficiency for large-scale deployments.

% \section*{CRediT authorship contribution statement}
% Junquan Gu: Conceptualization, Methodology, Software, Validation, Formal analysis, Investigation, Data curation, Visualization, Writing -- original draft, Writing -- review and editing. 

% Shibo Cui: Software, Validation, Formal analysis, Data curation, Visualization. 

% Xiangfeng Luo: Supervision, Methodology, Project administration, Writing -- review and editing. 

% Hang Yu: Supervision, Project administration, Resources, Writing -- review and editing.

% \section*{Declaration of competing interest}
% The authors declare that they have no known competing financial interests or personal relationships that could have appeared to influence the work reported in this paper.

% \section*{Data availability}
% The datasets used in this study are publicly available from Kaggle, UCI, and OpenML. The processed data splits, implementation details, and supplementary code will be made available at \url{https://github.com/sethGu/SAGE-Loop} upon publication.

% \section*{Funding}
% This research did not receive any specific grant from funding agencies in the public, commercial, or not-for-profit sectors.

\bibliographystyle{cas-model2-names}

\bibliography{sageloop}

@inproceedings{NIPS2015_11d0e628,
 author = {Feurer, Matthias and Klein, Aaron and Eggensperger, Katharina and Springenberg, Jost and Blum, Manuel and Hutter, Frank},
 booktitle = {Advances in Neural Information Processing Systems},
 editor = {C. Cortes and N. Lawrence and D. Lee and M. Sugiyama and R. Garnett},
 pages = {},
 publisher = {Curran Associates, Inc.},
 title = {Efficient and Robust Automated Machine Learning},
 url = {https://proceedings.neurips.cc/paper_files/paper/2015/file/11d0e6287202fced83f79975ec59a3a6-Paper.pdf},
 volume = {28},
 year = {2015}
}

@inproceedings{MLSYS2021_1ccc3bfa,
 author = {Wang, Chi and Wu, Qingyun and Weimer, Markus and Zhu, Erkang},
 booktitle = {Proceedings of Machine Learning and Systems},
 editor = {A. Smola and A. Dimakis and I. Stoica},
 pages = {434--447},
 title = {FLAML: A Fast and Lightweight AutoML Library},
 url = {https://proceedings.mlsys.org/paper_files/paper/2021/file/1ccc3bfa05cb37b917068778f3c4523a-Paper.pdf},
 volume = {3},
 year = {2021}
}

@inproceedings{
hollmann2022tabpfn,
title={Tab{PFN}: A Transformer That Solves Small Tabular Classification Problems in a Second},
author={Noah Hollmann and Samuel M{\"u}ller and Katharina Eggensperger and Frank Hutter},
booktitle={NeurIPS 2022 First Table Representation Workshop},
year={2022},
url={https://openreview.net/forum?id=eu9fVjVasr4}
}

@ARTICLE{Hollmann2025-xh,
  title    = "Accurate predictions on small data with a tabular foundation model",
  author   = "Hollmann, Noah and M{\"u}ller, Samuel and Purucker, Lennart and
              Krishnakumar, Arjun and K{\"o}rfer, Max and Hoo, Shi Bin and
              Schirrmeister, Robin Tibor and Hutter, Frank",
  journal  = "Nature",
  volume   =  637,
  number   =  8045,
  pages    = "319--326",
  month    =  jan,
  year     =  2025
}

@inproceedings{
hollmann2023large,
title={Large Language Models for Automated Data Science: Introducing {CAAFE} for Context-Aware Automated Feature Engineering},
author={Noah Hollmann and Samuel M{\"u}ller and Frank Hutter},
booktitle={Thirty-seventh Conference on Neural Information Processing Systems},
year={2023},
url={https://openreview.net/forum?id=9WSxQZ9mG7}
}

@article{
fang2024large,
title={Large Language Models ({LLM}s) on Tabular Data: Prediction, Generation, and Understanding - A Survey},
author={Xi Fang and Weijie Xu and Fiona Anting Tan and Ziqing Hu and Jiani Zhang and Yanjun Qi and Srinivasan H. Sengamedu and Christos Faloutsos},
journal={Transactions on Machine Learning Research},
issn={2835-8856},
year={2024},
url={https://openreview.net/forum?id=IZnrCGF9WI},
note={}
}

@InProceedings{pmlr-v220-roberts23a,
  title = 	 {AutoML Decathlon: Diverse Tasks, Modern Methods, and Efficiency at Scale},
  author =       {Roberts, Nicholas and Guo, Samuel and Xu, Cong and Talwalkar, Ameet and Lander, David and Tao, Lvfang and Cai, Linhang and Niu, Shuaicheng and Heng, Jianyu and Qin, Hongyang and Deng, Minwen and Hog, Johannes and Pfefferle, Alexander and Shivakumar, Sushil Ammanaghatta and Krishnakumar, Arjun and Wang, Yubo and Sukthanker, Rhea and Hutter, Frank and Hasanaj, Euxhen and Le, Tien-Dung and Khodak, Mikhail and Nevmyvaka, Yuriy and Rasul, Kashif and Sala, Frederic and Schneider, Anderson and Shen, Junhong and Sparks, Evan},
  booktitle = 	 {Proceedings of the NeurIPS 2022 Competitions Track},
  pages = 	 {151--170},
  year = 	 {2022},
  editor = 	 {Ciccone, Marco and Stolovitzky, Gustavo and Albrecht, Jacob},
  volume = 	 {220},
  series = 	 {Proceedings of Machine Learning Research},
  month = 	 {28 Nov--09 Dec},
  publisher =    {PMLR},
  url = 	 {https://proceedings.mlr.press/v220/roberts23a.html}
}

@misc{rozière2024codellamaopenfoundation,
      title={Code Llama: Open Foundation Models for Code}, 
      author={Baptiste Rozière and Jonas Gehring and Fabian Gloeckle and Sten Sootla and Itai Gat and Xiaoqing Ellen Tan and Yossi Adi and Jingyu Liu and Romain Sauvestre and Tal Remez and Jérémy Rapin and Artyom Kozhevnikov and Ivan Evtimov and Joanna Bitton and Manish Bhatt and Cristian Canton Ferrer and Aaron Grattafiori and Wenhan Xiong and Alexandre Défossez and Jade Copet and Faisal Azhar and Hugo Touvron and Louis Martin and Nicolas Usunier and Thomas Scialom and Gabriel Synnaeve},
      year={2024},
      eprint={2308.12950},
      archivePrefix={arXiv},
      primaryClass={cs.CL},
      url={https://arxiv.org/abs/2308.12950}, 
}

@misc{touvron2023llama2openfoundation,
      title={Llama 2: Open Foundation and Fine-Tuned Chat Models}, 
      author={Hugo Touvron and Louis Martin and Kevin Stone and Peter Albert and Amjad Almahairi and Yasmine Babaei and Nikolay Bashlykov and Soumya Batra and Prajjwal Bhargava and Shruti Bhosale and Dan Bikel and Lukas Blecher and Cristian Canton Ferrer and Moya Chen and Guillem Cucurull and David Esiobu and Jude Fernandes and Jeremy Fu and Wenyin Fu and Brian Fuller and Cynthia Gao and Vedanuj Goswami and Naman Goyal and Anthony Hartshorn and Saghar Hosseini and Rui Hou and Hakan Inan and Marcin Kardas and Viktor Kerkez and Madian Khabsa and Isabel Kloumann and Artem Korenev and Punit Singh Koura and Marie-Anne Lachaux and Thibaut Lavril and Jenya Lee and Diana Liskovich and Yinghai Lu and Yuning Mao and Xavier Martinet and Todor Mihaylov and Pushkar Mishra and Igor Molybog and Yixin Nie and Andrew Poulton and Jeremy Reizenstein and Rashi Rungta and Kalyan Saladi and Alan Schelten and Ruan Silva and Eric Michael Smith and Ranjan Subramanian and Xiaoqing Ellen Tan and Binh Tang and Ross Taylor and Adina Williams and Jian Xiang Kuan and Puxin Xu and Zheng Yan and Iliyan Zarov and Yuchen Zhang and Angela Fan and Melanie Kambadur and Sharan Narang and Aurelien Rodriguez and Robert Stojnic and Sergey Edunov and Thomas Scialom},
      year={2023},
      eprint={2307.09288},
      archivePrefix={arXiv},
      primaryClass={cs.CL},
      url={https://arxiv.org/abs/2307.09288}, 
}

@inproceedings{NEURIPS2023_43e9d647,
 author = {Liu, Jiawei and Xia, Chunqiu Steven and Wang, Yuyao and ZHANG, LINGMING},
 booktitle = {Advances in Neural Information Processing Systems},
 editor = {A. Oh and T. Naumann and A. Globerson and K. Saenko and M. Hardt and S. Levine},
 pages = {21558--21572},
 publisher = {Curran Associates, Inc.},
 title = {Is Your Code Generated by ChatGPT Really Correct? Rigorous Evaluation of Large Language Models for Code Generation},
 url = {https://proceedings.neurips.cc/paper_files/paper/2023/file/43e9d647ccd3e4b7b5baab53f0368686-Paper-Conference.pdf},
 volume = {36},
 year = {2023}
}

@inproceedings{10.1145/3551349.3559548,
author = {Khan, Junaed Younus and Uddin, Gias},
title = {Automatic Code Documentation Generation Using GPT-3},
year = {2023},
isbn = {9781450394758},
publisher = {Association for Computing Machinery},
address = {New York, NY, USA},
url = {https://doi.org/10.1145/3551349.3559548},
doi = {10.1145/3551349.3559548},
booktitle = {Proceedings of the 37th IEEE/ACM International Conference on Automated Software Engineering},
articleno = {174},
numpages = {6},
location = {Rochester, MI, USA},
series = {ASE '22}
}

@inproceedings{NEURIPS2020_62d75fb2,
 author = {Fakoor, Rasool and Mueller, Jonas W and Erickson, Nick and Chaudhari, Pratik and Smola, Alexander J},
 booktitle = {Advances in Neural Information Processing Systems},
 editor = {H. Larochelle and M. Ranzato and R. Hadsell and M.F. Balcan and H. Lin},
 pages = {8671--8681},
 publisher = {Curran Associates, Inc.},
 title = {Fast, Accurate, and Simple Models for Tabular Data via Augmented Distillation},
 url = {https://proceedings.neurips.cc/paper_files/paper/2020/file/62d75fb2e3075506e8837d8f55021ab1-Paper.pdf},
 volume = {33},
 year = {2020}
}

@misc{bubeck2023sparksartificialgeneralintelligence,
      title={Sparks of Artificial General Intelligence: Early experiments with GPT-4}, 
      author={Sébastien Bubeck and Varun Chandrasekaran and Ronen Eldan and Johannes Gehrke and Eric Horvitz and Ece Kamar and Peter Lee and Yin Tat Lee and Yuanzhi Li and Scott Lundberg and Harsha Nori and Hamid Palangi and Marco Tulio Ribeiro and Yi Zhang},
      year={2023},
      eprint={2303.12712},
      archivePrefix={arXiv},
      primaryClass={cs.CL},
      url={https://arxiv.org/abs/2303.12712}, 
}

@InProceedings{pmlr-v64-olson_tpot_2016,
  title = 	 {TPOT: A Tree-based Pipeline Optimization Tool for Automating Machine Learning},
  author = 	 {Olson, Randal S. and Moore, Jason H.},
  booktitle = 	 {Proceedings of the Workshop on Automatic Machine Learning},
  pages = 	 {66--74},
  year = 	 {2016},
  editor = 	 {Hutter, Frank and Kotthoff, Lars and Vanschoren, Joaquin},
  volume = 	 {64},
  series = 	 {Proceedings of Machine Learning Research},
  address = 	 {New York, New York, USA},
  month = 	 {24 Jun},
  publisher =    {PMLR},
  url = 	 {https://proceedings.mlr.press/v64/olson_tpot_2016.html}
}

@article{HE2021106622,
title = {AutoML: A survey of the state-of-the-art},
journal = {Knowledge-Based Systems},
volume = {212},
pages = {106622},
year = {2021},
issn = {0950-7051},
doi = {https://doi.org/10.1016/j.knosys.2020.106622},
url = {https://www.sciencedirect.com/science/article/pii/S0950705120307516},
author = {Xin He and Kaiyong Zhao and Xiaowen Chu}
}

@ARTICLE{Barbudo2023-uv,
  title    = "Eight years of {AutoML}: categorisation, review and trends",
  author   = "Barbudo, Rafael and Ventura, Sebasti{\'a}n and Romero, Jos{\'e}
              Ra{\'u}l",
  journal  = "Knowledge and Information Systems",
  volume   =  65,
  number   =  12,
  pages    = "5097--5149",
  month    =  dec,
  year     =  2023
}

@article{JMLR:v25:22-0493,
  author  = {Pieter Gijsbers and Marcos L. P. Bueno and Stefan Coors and Erin LeDell and S{{\'e}}bastien Poirier and Janek Thomas and Bernd Bischl and Joaquin Vanschoren},
  title   = {AMLB: an AutoML Benchmark},
  journal = {Journal of Machine Learning Research},
  year    = {2024},
  volume  = {25},
  number  = {101},
  pages   = {1--65},
  url     = {http://jmlr.org/papers/v25/22-0493.html}
}

@article{
xu2024large,
title={Large Language Models Synergize with Automated Machine Learning},
author={Jinglue Xu and Jialong Li and Zhen Liu and NAV Suryanarayanan and Guoyuan Zhou and JIA GUO and Hitoshi Iba and Kenji Tei},
journal={Transactions on Machine Learning Research},
issn={2835-8856},
year={2024},
url={https://openreview.net/forum?id=RDEaIfOiJM},
note={}
}

@article{
tornede2024automl,
title={Auto{ML} in the Age of Large Language Models: Current Challenges, Future Opportunities and Risks},
author={Alexander Tornede and Difan Deng and Theresa Eimer and Joseph Giovanelli and Aditya Mohan and Tim Ruhkopf and Sarah Segel and Daphne Theodorakopoulos and Tanja Tornede and Henning Wachsmuth and Marius Lindauer},
journal={Transactions on Machine Learning Research},
issn={2835-8856},
year={2024},
url={https://openreview.net/forum?id=cAthubStyG},
note={}
}

@misc{trirat2024automlagentmultiagentllmframework,
      title={AutoML-Agent: A Multi-Agent LLM Framework for Full-Pipeline AutoML}, 
      author={Patara Trirat and Wonyong Jeong and Sung Ju Hwang},
      year={2024},
      eprint={2410.02958},
      archivePrefix={arXiv},
      primaryClass={cs.LG},
      url={https://arxiv.org/abs/2410.02958}, 
}

@misc{zhang2023automlgptautomaticmachinelearning,
      title={AutoML-GPT: Automatic Machine Learning with GPT}, 
      author={Shujian Zhang and Chengyue Gong and Lemeng Wu and Xingchao Liu and Mingyuan Zhou},
      year={2023},
      eprint={2305.02499},
      archivePrefix={arXiv},
      primaryClass={cs.CL},
      url={https://arxiv.org/abs/2305.02499}, 
}

@misc{zheng2025opencodeinterpreterintegratingcodegeneration,
      title={OpenCodeInterpreter: Integrating Code Generation with Execution and Refinement}, 
      author={Tianyu Zheng and Ge Zhang and Tianhao Shen and Xueling Liu and Bill Yuchen Lin and Jie Fu and Wenhu Chen and Xiang Yue},
      year={2025},
      eprint={2402.14658},
      archivePrefix={arXiv},
      primaryClass={cs.SE},
      url={https://arxiv.org/abs/2402.14658}, 
}

@misc{lindauer2024positionactionhumancenteredautoml,
      title={Position: A Call to Action for a Human-Centered AutoML Paradigm}, 
      author={Marius Lindauer and Florian Karl and Anne Klier and Julia Moosbauer and Alexander Tornede and Andreas Mueller and Frank Hutter and Matthias Feurer and Bernd Bischl},
      year={2024},
      eprint={2406.03348},
      archivePrefix={arXiv},
      primaryClass={cs.LG},
      url={https://arxiv.org/abs/2406.03348}, 
}

@ARTICLE{Breiman2001-in,
  title    = "Random Forests",
  author   = "Breiman, Leo",
  journal  = "Machine Learning",
  volume   =  45,
  number   =  1,
  pages    = "5--32",
  month    =  oct,
  year     =  2001
}

@article{YU2025113809,
title = {Automated graph anomaly detection with large language models},
journal = {Knowledge-Based Systems},
volume = {324},
pages = {113809},
year = {2025},
issn = {0950-7051},
doi = {https://doi.org/10.1016/j.knosys.2025.113809},
url = {https://www.sciencedirect.com/science/article/pii/S095070512500855X},
author = {Jiaqi Yu and Yang Gao and Hong Yang and Zhihong Tian and Peng Zhang and Xingquan Zhu}
}

@Article{computers10020024,
AUTHOR = {Mustafa, Akram and Rahimi Azghadi, Mostafa},
TITLE = {Automated Machine Learning for Healthcare and Clinical Notes Analysis},
JOURNAL = {Computers},
VOLUME = {10},
YEAR = {2021},
NUMBER = {2},
ARTICLE-NUMBER = {24},
URL = {https://www.mdpi.com/2073-431X/10/2/24},
ISSN = {2073-431X},
DOI = {10.3390/computers10020024}
}

@INPROCEEDINGS{9002839,
  author={Wang, Can and Bäck, Thomas and Hoos, Holger H. and Baratchi, Mitra and Limmer, Steffen and Olhofer, Markus},
  booktitle={2019 IEEE Symposium Series on Computational Intelligence (SSCI)}, 
  title={Automated Machine Learning for Short-term Electric Load Forecasting}, 
  year={2019},
  volume={},
  number={},
  pages={314-321},
  doi={10.1109/SSCI44817.2019.9002839}}

@article{DBLP:journals/corr/abs-2111-02508,
  author       = {Iddo Drori and
                  Yamuna Krishnamurthy and
                  R{\'{e}}mi Rampin and
                  Raoni de Paula Louren{\c{c}}o and
                  Jorge Piazentin Ono and
                  Kyunghyun Cho and
                  Cl{\'{a}}udio T. Silva and
                  Juliana Freire},
  title        = {AlphaD3M: Machine Learning Pipeline Synthesis},
  journal      = {CoRR},
  volume       = {abs/2111.02508},
  year         = {2021},
  url          = {https://arxiv.org/abs/2111.02508},
  eprinttype    = {arXiv},
  eprint       = {2111.02508},
  bibsource    = {dblp computer science bibliography, https://dblp.org}
}

@article{NIKITIN2024112363,
title = {Integration of evolutionary automated machine learning with structural sensitivity analysis for composite pipelines},
journal = {Knowledge-Based Systems},
volume = {302},
pages = {112363},
year = {2024},
issn = {0950-7051},
doi = {https://doi.org/10.1016/j.knosys.2024.112363},
url = {https://www.sciencedirect.com/science/article/pii/S0950705124009973},
author = {Nikolay O. Nikitin and Maiia Pinchuk and Valerii Pokrovskii and Peter Shevchenko and Andrey Getmanov and Yaroslav Aksenkin and Ilia Revin and Andrey Stebenkov and Vladimir Latypov and Ekaterina Poslavskaya and Anna V. Kalyuzhnaya}
}

@inproceedings{10.1145/3664647.3680665,
author = {Luo, Daqin and Feng, Chengjian and Nong, Yuxuan and Shen, Yiqing},
title = {AutoM3L: An Automated Multimodal Machine Learning Framework with Large Language Models},
year = {2024},
isbn = {9798400706868},
publisher = {Association for Computing Machinery},
address = {New York, NY, USA},
url = {https://doi.org/10.1145/3664647.3680665},
doi = {10.1145/3664647.3680665},
booktitle = {Proceedings of the 32nd ACM International Conference on Multimedia},
pages = {8586–8594},
numpages = {9},
location = {Melbourne VIC, Australia},
series = {MM '24}
}

@inproceedings{olson2016tpot,
  title={TPOT: A tree-based pipeline optimization tool for automating machine learning},
  author={Olson, Randal S and Moore, Jason H},
  booktitle={Workshop on automatic machine learning},
  pages={66--74},
  year={2016},
  organization={PMLR}
}

@article{erickson2020autogluon,
  title={Autogluon-tabular: Robust and accurate automl for structured data},
  author={Erickson, Nick and Mueller, Jonas and Shirkov, Alexander and Zhang, Hang and Larroy, Pedro and Li, Mu and Smola, Alexander},
  journal={arXiv preprint arXiv:2003.06505},
  year={2020}
}

@inproceedings{ledell2020h2o,
  title={H2o automl: Scalable automatic machine learning},
  author={LeDell, Erin and Poirier, Sebastien},
  booktitle={Proceedings of the AutoML Workshop at ICML},
  volume={2020},
  pages={24},
  year={2020}
}

@inproceedings{chen2016xgboost,
  title={Xgboost: A scalable tree boosting system},
  author={Chen, Tianqi and Guestrin, Carlos},
  booktitle={Proceedings of the 22nd acm sigkdd international conference on knowledge discovery and data mining},
  pages={785--794},
  year={2016}
}

@article{ke2017lightgbm,
  title={Lightgbm: A highly efficient gradient boosting decision tree},
  author={Ke, Guolin and Meng, Qi and Finley, Thomas and Wang, Taifeng and Chen, Wei and Ma, Weidong and Ye, Qiwei and Liu, Tie-Yan},
  journal={Advances in neural information processing systems},
  volume={30},
  year={2017}
}

@article{guo2024ds,
  title={Ds-agent: Automated data science by empowering large language models with case-based reasoning},
  author={Guo, Siyuan and Deng, Cheng and Wen, Ying and Chen, Hechang and Chang, Yi and Wang, Jun},
  journal={arXiv preprint arXiv:2402.17453},
  year={2024}
}

@article{10.5555/3586589.3586850,
author = {Feurer, Matthias and Eggensperger, Katharina and Falkner, Stefan and Lindauer, Marius and Hutter, Frank},
title = {Auto-sklearn 2.0: hands-free AutoML via meta-learning},
year = {2022},
issue_date = {January 2022},
publisher = {JMLR.org},
volume = {23},
number = {1},
issn = {1532-4435},
journal = {J. Mach. Learn. Res.},
month = jan,
articleno = {261},
numpages = {61}
}

@misc{
tsymbalov2024llmfeatures,
title={{LLM}2Features: Large Language Models in Interpretable Feature Generation for Auto{ML} with Tabular Data},
author={Aleksandr Tsymbalov and Andrey Savchenko},
year={2024},
url={https://openreview.net/forum?id=qbSoiHLEK0}
}

@article{DBLP:journals/corr/abs-2012-05390,
  author       = {Jason Yoo and
                  Tony Joseph and
                  Dylan Yung and
                  Seyed Ali Nasseri and
                  Frank Wood},
  title        = {Ensemble Squared: {A} Meta AutoML System},
  journal      = {CoRR},
  volume       = {abs/2012.05390},
  year         = {2020},
  url          = {https://arxiv.org/abs/2012.05390},
  eprinttype    = {arXiv},
  eprint       = {2012.05390},
  bibsource    = {dblp computer science bibliography, https://dblp.org}
}

@InProceedings{pmlr-v224-purucker23b,
  title = 	 {Q(D)O-ES: Population-based Quality (Diversity) Optimisation for Post Hoc Ensemble Selection in AutoML},
  author =       {Purucker, Lennart Oswald and Schneider, Lennart and Anastacio, Marie and Beel, Joeran and Bischl, Bernd and Hoos, Holger},
  booktitle = 	 {Proceedings of the Second International Conference on Automated Machine Learning},
  pages = 	 {10/1--34},
  year = 	 {2023},
  editor = 	 {Faust, Aleksandra and Garnett, Roman and White, Colin and Hutter, Frank and Gardner, Jacob R.},
  volume = 	 {224},
  series = 	 {Proceedings of Machine Learning Research},
  month = 	 {12--15 Nov},
  publisher =    {PMLR},
  url = 	 {https://proceedings.mlr.press/v224/purucker23b.html}
}

@article{10.1145/3643564,
author = {Poulakis, Yannis and Doulkeridis, Christos and Kyriazis, Dimosthenis},
title = {A Survey on AutoML Methods and Systems for Clustering},
year = {2024},
issue_date = {June 2024},
publisher = {Association for Computing Machinery},
address = {New York, NY, USA},
volume = {18},
number = {5},
issn = {1556-4681},
url = {https://doi.org/10.1145/3643564},
doi = {10.1145/3643564},
journal = {ACM Trans. Knowl. Discov. Data},
month = feb,
articleno = {120},
numpages = {30}
}

@Article{app15126876,
AUTHOR = {Schlake, Georg Stefan and Pernklau, Max and Beecks, Christian},
TITLE = {Automated Exploratory Clustering to Democratize Clustering Analysis},
JOURNAL = {Applied Sciences},
VOLUME = {15},
YEAR = {2025},
NUMBER = {12},
ARTICLE-NUMBER = {6876},
URL = {https://www.mdpi.com/2076-3417/15/12/6876},
ISSN = {2076-3417},
DOI = {10.3390/app15126876}
}

@article{10.1162/153244303321897735,
author = {Strehl, Alexander and Ghosh, Joydeep},
title = {Cluster ensembles --- a knowledge reuse framework for combining multiple partitions},
year = {2003},
issue_date = {3/1/2003},
publisher = {JMLR.org},
volume = {3},
number = {null},
issn = {1532-4435},
url = {https://doi.org/10.1162/153244303321897735},
doi = {10.1162/153244303321897735},
journal = {J. Mach. Learn. Res.},
month = mar,
pages = {583–617},
numpages = {35}
}

@article{SAHIN2023105732,
title = {Greedy-AutoML: A novel greedy-based stacking ensemble learning framework for assessing soil liquefaction potential},
journal = {Engineering Applications of Artificial Intelligence},
volume = {119},
pages = {105732},
year = {2023},
issn = {0952-1976},
doi = {https://doi.org/10.1016/j.engappai.2022.105732},
url = {https://www.sciencedirect.com/science/article/pii/S0952197622007229},
author = {Emrehan Kutlug Sahin and Selcuk Demir}
}

@ARTICLE{Rimal2024-hx,
  title    = "Machine learning model matters its accuracy: a comparative study
              of ensemble learning and {AutoML} using heart disease prediction",
  author   = "Rimal, Yagyanath and Paudel, Siddhartha and Sharma, Navneet and
              Alsadoon, Abeer",
  journal  = "Multimedia Tools and Applications",
  volume   =  83,
  number   =  12,
  pages    = "35025--35042",
  month    =  apr,
  year     =  2024
}

@ARTICLE{9034023,
  author={Yu, Hang and Lu, Jie and Zhang, Guangquan},
  journal={IEEE Transactions on Knowledge and Data Engineering}, 
  title={An Online Robust Support Vector Regression for Data Streams}, 
  year={2022},
  volume={34},
  number={1},
  pages={150-163},
  doi={10.1109/TKDE.2020.2979967}}

@ARTICLE{8897716,
  author={Yu, Hang and Lu, Jie and Zhang, Guangquan},
  journal={IEEE Transactions on Neural Networks and Learning Systems}, 
  title={Online Topology Learning by a Gaussian Membership-Based Self-Organizing Incremental Neural Network}, 
  year={2020},
  volume={31},
  number={10},
  pages={3947-3961},
  doi={10.1109/TNNLS.2019.2947658}}

@ARTICLE{9194301,
  author={Yu, Hang and Lu, Jie and Zhang, Guangquan},
  journal={IEEE Transactions on Cybernetics}, 
  title={Continuous Support Vector Regression for Nonstationary Streaming Data}, 
  year={2022},
  volume={52},
  number={5},
  pages={3592-3605},
  doi={10.1109/TCYB.2020.3015266}}

@article{WEN2024111636,
title = {Adaptive tree-like neural network: Overcoming catastrophic forgetting to classify streaming data with concept drifts},
journal = {Knowledge-Based Systems},
volume = {293},
pages = {111636},
year = {2024},
issn = {0950-7051},
doi = {https://doi.org/10.1016/j.knosys.2024.111636},
url = {https://www.sciencedirect.com/science/article/pii/S0950705124002715},
author = {YiMin Wen and Xiang Liu and Hang Yu}
}

@inproceedings{10.1145/3696410.3714567,
author = {Liu, Zhengyang and Yu, Hang and Luo, Xiangfeng},
title = {Federated Graph Anomaly Detection via Disentangled Representation Learning},
year = {2025},
isbn = {9798400712746},
publisher = {Association for Computing Machinery},
address = {New York, NY, USA},
url = {https://doi.org/10.1145/3696410.3714567},
doi = {10.1145/3696410.3714567},
booktitle = {Proceedings of the ACM on Web Conference 2025},
pages = {1216–1224},
numpages = {9},
location = {Sydney NSW, Australia},
series = {WWW '25}
}

@article{Yu_Liu_Luo_2024, title={Barely Supervised Learning for Graph-Based Fraud Detection}, volume={38}, url={https://ojs.aaai.org/index.php/AAAI/article/view/29593}, DOI={10.1609/aaai.v38i15.29593}, abstractNote={In recent years, graph-based fraud detection methods have garnered increasing attention for their superior ability to tackle the issue of camouflage in fraudulent scenarios. However, these methods often rely on a substantial proportion of samples as the training set, disregarding the reality of scarce annotated samples in real-life scenarios. As a theoretical framework within semi-supervised learning, the principle of consistency regularization posits that unlabeled samples should be classified into the same category as their own perturbations. Inspired by this principle, this study incorporates unlabeled samples as an auxiliary during model training, designing a novel barely supervised learning method to address the challenge of limited annotated samples in fraud detection. Specifically, to tackle the issue of camouflage in fraudulent scenarios, we employ disentangled representation learning based on edge information for a small subset of annotated nodes. This approach partitions node features into three distinct components representing different connected edges, providing a foundation for the subsequent augmentation of unlabeled samples. For the unlabeled nodes used in auxiliary training, we apply both strong and weak augmentation and design regularization losses to enhance the detection performance of the model in the context of extremely limited labeled samples. Across five publicly available datasets, the proposed model showcases its superior detection capability over baseline models.}, number={15}, journal={Proceedings of the AAAI Conference on Artificial Intelligence}, author={Yu, Hang and Liu, Zhengyang and Luo, Xiangfeng}, year={2024}, month={Mar.}, pages={16548-16557} }

@inproceedings{10.1145/3701716.3715587,
author = {Gu, Junquan and Yu, Hang and Luo, Xiangfeng},
title = {A Masked AutoEncoder with Strong-Weak Mutual Information for Anomaly Detection in Dynamic Incomplete Graphs},
year = {2025},
isbn = {9798400713316},
publisher = {Association for Computing Machinery},
address = {New York, NY, USA},
url = {https://doi.org/10.1145/3701716.3715587},
doi = {10.1145/3701716.3715587},
booktitle = {Companion Proceedings of the ACM on Web Conference 2025},
pages = {986–990},
numpages = {5},
location = {Sydney NSW, Australia},
series = {WWW '25}
}

@INPROCEEDINGS{6119056,
  author={Jiang, Jing and Lu, Jie and Zhang, Guangquan},
  booktitle={2011 IEEE Ninth International Conference on Dependable, Autonomic and Secure Computing}, 
  title={An Innovative Self-Adaptive Configuration Optimization System in Cloud Computing}, 
  year={2011},
  volume={},
  number={},
  pages={621-627},
  doi={10.1109/DASC.2011.112}}

@InProceedings{10.1007/11875581_6,
author="Hao, Zhifeng
and Wen, Wen
and Yang, Xiaowei
and Lu, Jie
and Zhang, Guangquan",
editor="Corchado, Emilio
and Yin, Hujun
and Botti, Vicente
and Fyfe, Colin",
title="A Fast Data Preprocessing Procedure for Support Vector Regression",
booktitle="Intelligent Data Engineering and Automated Learning -- IDEAL 2006",
year="2006",
publisher="Springer Berlin Heidelberg",
address="Berlin, Heidelberg",
pages="48--56",
isbn="978-3-540-45487-8"
}


% Author biography without author photo.
% Author biography. Author biography. Author biography.
% Author biography. Author biography. Author biography.
% Author biography. Author biography. Author biography.
% Author biography. Author biography. Author biography.
% Author biography. Author biography. Author biography.
% Author biography. Author biography. Author biography.
% Author biography. Author biography. Author biography.
% Author biography. Author biography. Author biography.
% Author biography. Author biography. Author biography.

% \bio{figs/cas-pic1}
% Author biography with author photo.
% Author biography. Author biography. Author biography.
% Author biography. Author biography. Author biography.
% Author biography. Author biography. Author biography.
% Author biography. Author biography. Author biography.
% Author biography. Author biography. Author biography.
% Author biography. Author biography. Author biography.
% Author biography. Author biography. Author biography.
% Author biography. Author biography. Author biography.
% Author biography. Author biography. Author biography.

% \bio{figs/cas-pic1}
% Author biography with author photo.
% Author biography. Author biography. Author biography.
% Author biography. Author biography. Author biography.
% Author biography. Author biography. Author biography.
% Author biography. Author biography. Author biography.

\end{document}